\documentclass[preprint,10pt]{elsarticle}

\usepackage{amsmath,amsthm,amssymb,amsfonts}
\usepackage{algorithmic}
\usepackage{algorithm}
\usepackage{array}
\usepackage{textcomp}
\usepackage{stfloats}
\usepackage{url}
\usepackage{verbatim}
\usepackage{graphicx}
\usepackage{multirow}
\usepackage{booktabs}
\usepackage{float}
\usepackage{subfigure}
\usepackage{makecell}
\usepackage{pifont}
\usepackage{bm}
\usepackage{rotating}
\usepackage{placeins}

\begin{document}

\title{TGFormer: Towards Temporal Graph Transformer with Auto-Correlation Mechanism}


\author[label1]{Hongjiang Chen}
\ead{hchen@hdu.edu.cn}

\author[label1]{Pengfei Jiao\corref{cor1}}
\ead{pjiao@hdu.edu.cn}

\author[label1]{Ming Du}
\ead{mdu@hdu.edu.cn}

\author[label2]{Xuan Guo}
\ead{guoxuan@tju.edu.cn}

\author[label1]{Zhidong Zhao}
\ead{zhaozd@hdu.edu.cn}
\author[label2]{Di~Jin}
\ead{jindi@tju.edu.cn}

\author[label3]{Xiao Liu}
\ead{lxc417@sjtu.edu.cn}

\cortext[cor1]{Corresponding author}

\affiliation[label1]{
            addressline={Hangzhou Dianzi University, School of Cyberspace}, 
            city={Hangzhou},
            postcode={310018},
            country={China}}
\affiliation[label2]{
            addressline={Tianjin University, College of Intelligence and Computing}, 
            city={Tianjin},
            postcode={300072},
            country={China}}
\affiliation[label3]{
            addressline={State Key Laboratory of Systems Medicine for Cancer, Shanghai Cancer Institute}, 
            city={Shanghai},
            postcode={200127},
            country={China}}


\begin{abstract}
The growing interest in Temporal Graph Neural Networks (TGNNs) stems from their ability to model complex dynamics and deliver superior performance. However, TGNNs encounter fundamental challenges in capturing long-term dependencies and identifying periodic patterns. To address these limitations, we propose TGFormer, a novel Transformer architecture specifically designed for temporal graphs. Our model redefines temporal graph learning by establishing a trajectory framework that aligns with time series analysis principles. This approach allows TGFormer to derive node representations through systematic analysis of historical interactions, enabling granular examination of node relationships across sequential timestamps. Building upon stochastic process theory, we develop an auto-correlation mechanism that systematically uncovers periodic dependencies in node interactions. This innovation empowers TGFormer to perform dependency discovery and representation aggregation at sub-interaction levels, demonstrating superior efficiency and accuracy compared to conventional attention mechanisms. Experimental validation across six public benchmarks confirms the effectiveness of our approach, with TGFormer at most achieving 9.35\% precision improvement compared to state-of-the-art approaches.
\end{abstract}




\maketitle

\section{Introduction}
Recent years have witnessed a significant increase in research focused on representation learning for temporal graphs~\cite{jiao2025survey}. This surge is driven by the applicability of temporal graphs in modeling dynamic systems such as social networks~\cite{wang2025continual}, traffic networks~\cite{weng2023decomposition}, and financial transaction systems~\cite{sun2025knowledge}. Temporal Graph Neural Networks (TGNNs) have emerged as a promising framework by effectively capturing complex temporal dependencies, thereby enabling significant improvements in tasks such as dynamic community detection~\cite{bai2024haqjsk}, traffic flow prediction~\cite{kumar2024tsanet}, and real-time recommendation systems~\cite{zhang2020social}. TGNN methodologies are primarily categorized into discrete-time temporal graph (DTTG) methods~\cite{jiao2024contrastive} and continuous-time temporal graph (CTTG) methods~\cite{chen2023temporal}. DTTG methods process graph snapshots at predetermined intervals, which may lead to information loss, whereas CTTG methods directly model event sequences to capture the finer nuances of temporal irregularities~\cite{qin2023temporal}. This clear distinction motivates our focus on advancing CTTG learning techniques.

Despite the successes of current CTTG methods, they encounter two critical challenges in modeling dynamic interactions. First, the popular memory-based and walk-based methods, such as TGN~\cite{rossi2020temporal} and CAWN~\cite{wang2021inductive}, rely on extensive historical data storage or sampling~\cite{souza2022provably}, which limits their ability to model long-term dependencies essential for applications like predicting user engagement trends in social networks over months or years. 
Second, most methods frequently fail to capture intrinsic periodic patterns that dominate many real-world systems~\cite{li2025hypergraph, li2024guest}. These patterns, such as e-commerce platforms exhibit weekly cycles with sales peaks on weekends while streaming services experience daily viewership surges during evening hours are crucial for forecasting future behavior. Transformer-based models like DyGFormer~\cite{yu2023towards} attempt to address these issues by extending historical context through attention mechanisms, as they naturally capture long-term dependencies. However, their standard attention design acts as a low-pass filter, averaging historical data and smoothing out high-frequency periodic signals~\cite{wu2021autoformer}. This behavior, rooted in the attention mechanism's tendency to assign diffuse weights across time steps, has been shown in empirical studies~\cite{zhao2025tfformer,dai2024ddn} to obscure sharp periodic fluctuations, such as weekend sales spikes, treating them as noise rather than predictive features. As a result, these methods yield suboptimal predictions in scenarios where periodic patterns are central to understanding temporal dynamics.

To address these challenges, we introduce TGFormer, a novel Transformer-based architecture designed for CTTGs, which efficiently captures long-term dependencies using a Transformer framework, leveraging its ability to process historical data effectively. To handle periodic patterns, TGFormer treats temporal graph learning as a time-series analysis task and employs an auto-correlation mechanism (ACoM) instead of traditional attention mechanisms. Drawing from stochastic process theory~\cite{cox2017theory}, ACoM transforms time series from the time domain to the frequency domain using the Fast Fourier Transform (FFT)~\cite{duhamel1990fast}, where periodic patterns are represented as distinct frequency components, enabling the model to focus on specific frequency bands. This frequency domain approach is crucial because standard attention mechanisms, operating in the time domain, tend to smooth out high-frequency signals by averaging historical data, treating periodic fluctuations as noise, and obscuring them. By contrast, ACoM's frequency domain operation preserves these fluctuations, enhancing the capture of periodic patterns while reducing computational complexity from \( O(L^2) \) to \( O(L \log L) \). Our extensive experiments on six real-world datasets demonstrate TGFormer's effectiveness, achieving at most up to 9.35\% improvement in precision compared to state-of-the-art methods, validating its ability to accurately model both long-term dependencies and periodic structures in temporal graphs.

The primary contributions of this work are summarized as follows:
\begin{itemize}
    \item Treating temporal graph learning as a time-series analysis problem is a novel perspective, enabling the application of time-series techniques to capture complex temporal patterns, including periodicities, which are crucial for temporal graph representation learning.
    \item Based on this perspective, we design the Series Transformer layer, a novel component that effectively captures long-term dependencies in temporal graphs by modeling historical data using Transformer-based architecture.
    \item We introduce the auto-correlation mechanism, a novel method that complements the Series Transformer layer by capturing periodic patterns through sequence auto-correlation, enhancing the model's ability to handle time series data with reduced computational complexity (\( O(L \log L) \)).
    \item We validate TGFormer's performance through comprehensive experiments on six real-world temporal graph datasets, showing that it consistently outperforms state-of-the-art methods, demonstrating its effectiveness in capturing both long-term dependencies and periodic structures.
\end{itemize}

\section{Related Work}
\begin{table}[!ht]
    \centering
    \caption{The Qualitative Comparison of the Current Method}
    \begin{tabular}{c|ccc}
        \toprule
        method & Data Type & Transformer & Period Pattern\\ \midrule
         Autoformer~\cite{wu2021autoformer} & Time Series & \ding{52} & \ding{52} \\
         FEDformer~\cite{zhou2022fedformer} & Time Series &  \ding{52} & \ding{56} \\
         SGFormer~\cite{wu2023sgformer} & Static Graph &  \ding{52} & \ding{56} \\
         Difformer~\cite{wu2023difformer} & Static Graph &  \ding{52} & \ding{56} \\
         GraphMixer~\cite{cong2023we} & Temporal Graph & \ding{56} & \ding{56} \\
         DyGFormer~\cite{yu2023towards} & Temporal Graph & \ding{52} & \ding{56} \\
         SimpleDyG~\cite{wu2024feasibility} & Temporal Graph & \ding{52} & \ding{56} \\ \midrule
         \textbf{TGFormer} & \textbf{Temporal Graph} & \ding{52} & \ding{52} \\ \bottomrule
    \end{tabular}
    \label{tab: Related}
\end{table}
\subsection{Temporal Graph Neural Networks}

In recent years, research on TGNNs has garnered increasing attention. These models typically conceptualize temporal graph data as event streams, enabling the direct learning of node representations from continuously evolving interactions. Specifically, existing CTTG models predominantly employ recurrent neural networks (RNNs) or attention mechanisms as their sequence modeling components. For instance, JODIE~\cite{kumar2019predicting} utilizes an RNN to integrate static and dynamic embeddings for node representation learning, while TGAT~\cite{xu2020inductive} extends the graph attention mechanism to capture time-aware representations. Building upon these foundations, several approaches incorporate additional techniques to enhance the modeling of continuous temporal dynamics. For example, TGN~\cite{rossi2020temporal}, a variant of TGAT, introduces a memory module to effectively track node-level feature evolution. GSNOP~\cite{luo2023graph} employs a sequential ordinary differential equation (ODE) aggregator, leveraging sequential dependencies to learn the derivative of neural processes, thereby improving distribution estimation from limited historical data. Simultaneously, CAWN~\cite{wang2021inductive} extracts temporal network motifs using set-based anonymous random walks, effectively capturing dynamic structural patterns. NeurTW~\cite{jin2022neural} further enhances spatial and temporal interdependencies during anonymous random walks, integrating continuous evolution with transient activation processes to model the underlying spatiotemporal dynamics of base-order coding. To address long-term temporal dependencies in temporal graph interactions, DyGFormer~\cite{yu2023towards} adopts a Transformer-based architecture, incorporating a neighbor co-occurrence encoding scheme alongside a patching technique to refine representation learning.

In a nutshell, most existing TGNNs struggle to manage nodes with longer interactions and effectively capture periodic temporal patterns due to the prohibitive computational costs of complex modules and optimization challenges such as vanishing or exploding gradients. In this paper, we propose a novel transformer tailored for temporal graphs to show the necessity of long-term temporal dependencies and periodic dependencies, which is achieved by the Series Transformer layer.


\subsection{Transformers on Graphs}
Transformer~\cite{vaswani2017attention} is an innovative model that employs the SAM to handle sequential data, which has demonstrated significant success across diverse domains. Recent advances in graph learning have begun to assimilate Transformer models in diverse formulations. For instance, NAGphormer~\cite{chen2023nagphormer} employs sampling-based node selection, while Difformer~\cite{wu2023difformer} proposes continuous-time diffusion for dynamic graphs. Exphormer~\cite{shirzad2023exphormer} enhances structural awareness through expander graph augmentation and virtual nodes, outperforming traditional sparse Transformers. NodeFormer~\cite{wu2022nodeformer} leverages Gumbel-Softmax kernels for all-pair message passing, and SGFormer~\cite{wu2023sgformer}  shows that just using a one-layer transformer network can sometimes improve the results of GCN-based networks and the low memory footprint can help scale to large networks. SimpleDyG~\cite{wu2024feasibility} re-conceptualizes temporal graphs as a sequence added to the attention mechanism.

Although some studies have explored the potential of transformers for temporal graphs, they have only enhanced the inputs to the transformer and have not proposed a transformer that can accommodate characteristics unique to temporal graphs, such as periodicity. Addressing this, our work architects an auto-correlation methodology to unearth the periodic dependencies of node interactions and undertake dependency discovery and representation aggregation at the sub-interaction level. 

\textbf{Remark.} For qualitative comparison, we summarize recent major temporal graph learning and transformer learning methods in Table~\ref{tab: Related}. The comparison encompasses three aspects: applicability to temporal Graph, adoption of Transformer learning manner, and consideration of period temporal pattern. To the best of our knowledge, our approach is the only one that simultaneously satisfies all three aspects.


\begin{figure*}[!t]
    \centering
    \vspace{-0.1in}
    \includegraphics[width=0.95\textwidth]{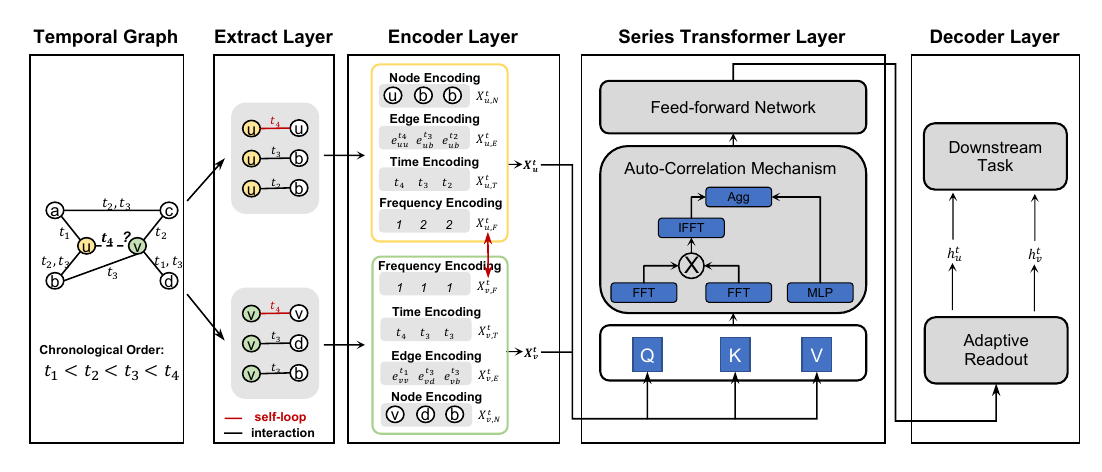}
    \caption{The overview of TGFormer begins with the extract layer, which employs a direct interaction extractor. Subsequently, the encoder layer generates temporal-aware and structure-aware sequence node representations. These representations are then used to form the time series data placed into the query (Q), key (K), and value (V) matrices in the series transformer layer. The series transformer layer leverages the Transformer's ability to capture long-term temporal dependencies and the auto-correlation mechanism to uncover inner periodic temporal patterns in the series data. Finally, the decoder layer utilizes adaptive readout for various downstream tasks.}
    \label{fig: model}
\end{figure*}

\begin{table}[!ht]
\caption{Symbols and their definitions}
\centering
{
\renewcommand\arraystretch{1.1}
\resizebox{\linewidth}{!}{
\begin{tabular}{cl}
\toprule
\bf{Notation} & \bf{Description}\\
\hline
  $\mathcal{G}$ & A temporal graph \\
  $\mathcal{V}$, $\mathcal{E}$ & The node set and link set of $\mathcal{G}$ \\
  $S^{t}_*$ & 
  \makecell[l]{The historical interactions samples of node $*$ before time $t$ with itself}\\
  $X^t_{*, N}$ & Node $*$'s representation of node features at time $t$ \\
  $X^t_{*, E}$ & Node $*$'s representation of link features at time $t$ \\
  $X^t_{*, T}$ & Node $*$'s representation of temporal information at time $t$ \\
  $X^t_{*, F}$ & Node $*$'s representation of node interactions frequency at time $t$ \\
  $F^t_*$ & The number of times a node appears in $S_u^t$ and $S_v^t$, respectively \\
  $X_*^t$ & The combine node $*$ representation at time $t$\\
  $h_*^t$ & The node $*$ representation at time $t$\\
  $d_N, d_E, d_T, d_F, d$ & The dimension of $X^t_{*, N}, X^t_{*, E}, X^t_{*, T}, X^t_{*, F}$ \\
  $\left\{\mathcal{X}^t_i\right\}_{i=1}^L$ & A time series \\
  $\delta$ & The time delay\\
  $k$ & The select k series\\ 
  $L$ & The node historical interactions extract length \\
\bottomrule
\end{tabular}}
}
 \label{tab: notation}
\end{table}
\section{Preliminaries}
\subsection{Temporal Graph}

A temporal graph is defined as $\mathcal{G} = (\mathcal{V}, \mathcal{E})$, where $\mathcal{V}$ represents the set of nodes and $\mathcal{E}$ denotes the set of node interactions, each associated with a timestamp. For any two nodes $u$ and $v$, there may exist a sequence of interactions occurring at different timestamps, formally expressed as $\mathcal{E}_{u, v} = \left\{(u, v, t_1), (u, v, t_2), \dots, (u, v, t_n)\right\} \subset \mathcal{E},$ where the timestamps are ordered as $0 < t_1 < t_2 < \dots < t_n.$ This sequence indicates that nodes $u$ and $v$ have interacted at least once at each corresponding timestamp. Two nodes that interact are referred to as neighbors. The symbols and their definitions are summarized in Table~\ref{tab: notation}, and this notation will be used in the subsequent sections.

\subsection{Temporal Link Prediction}
We aim to learn a model that given a pair of nodes with a specific timestamp $t$, we aim to predict whether the two nodes are connected at $t$ based on all the available historical data. Note that we are not only concerned with the prediction of links between nodes seen during training. We also expect the model to predict links between nodes that are never seen for inductive evaluation. 
For a more reliable comparison, we use the three strategies in~\cite{poursafaei2022towards} (i.e., random, historical, and inductive negative sampling strategies) to comprehensively evaluate the performance of the model on the temporal link prediction task.

\subsection{Transformer}
A Transformer block relies heavily on a multi-head attention mechanism to learn the context-aware representation for sequences, which can be defined as:
\begin{equation}
\begin{aligned}
O=\operatorname{MultiHead}(Q,K,V) =W_O  \operatorname{Concat}(O_1,O_2,\cdots,O_{h}),
\end{aligned}
\end{equation}
where $W_O$ is a trainable parameter. $h$ is the number of heads. $O_i$is computed as:
\begin{equation}
O_i = \operatorname{Attention}(Q_i,K_i,V_i) = \operatorname{SoftMax}\left(\frac{Q_i K_i^T}{\sqrt{d}}\right)V_i,
\end{equation}
where $\operatorname{Attention}(\cdot, \cdot, \cdot)$ is the scaled dot-product attention.
As input to the Transformer, an element in a sequence is represented by an embedding vector. The multi-head attention mechanism works by injecting the Positional Embedding into the element embeddings to be aware of element orders in a sequence.

\section{Proposed Method}
The framework of our TGFormer is shown in Fig.~\ref{fig: model}, which employs a Series Transformer as the backbone. Given an interaction $(u, v, t)$, we first extract historical interactions of source node $u$ and destination node $v$ before timestamp $t$ and obtain two interaction sequences $S^t_u$ and $S^t_v$. Next, computing the encodings of neighbors, links, time intervals, and interaction frequency for each sequence. Then, we concat each encoding sequence feed into the Series Transformer for capturing long-term temporal dependencies and periodic dependencies. Finally, the outputs of the Transformer adopt an adaptive readout function to derive time-aware representations of $u$ and $v$ at timestamp $t$ (i.e., $h^t_u$ and $h^t_v$), which can be applied in various downstream tasks like temporal link prediction.
\subsection{Extract Layer}
In our initial endeavor of predicting the connection between the nodal elements $u$ and $v$ at a specific timestamp $t$, we first affix each node with a self-loop at the specific time $t$. This maneuver aims to enhance the correlation of the node to its inherent characteristics.
Subsequently, we extract $L$ interactions for both nodes, ordered based on temporal proximity, which could interact with the same node at different timestamps. If a node has fewer than $L$ historical neighbors, zero-padding is employed to reconcile the deficit.
This process culminates in the successful transmutation of the intricate problem of temporal graph learning into a more comprehensible paradigm - a time series analysis problem.
Mathematically, given the predictive interaction $(u, v, t)$, for the source node $u$ and the target node $v$, we exact the series of interactions encompassing nodes $u$ and $v$, expressed as $S_u^t = \left\{ (u, u^\prime, t^\prime) | t^\prime < t\right\} \cup \left\{ (u, u, t)\right\}$ and $S_v^t = \left\{ (v, v^\prime, t^\prime) | t^\prime < t\right\} \cup \left\{ (v, v, t)\right\}$, respectively. Note, $(u,u^\prime, t^\prime) \in \mathcal{E}, (v,v^\prime, t^\prime) \in \mathcal{E}$, and $ |S_u^t|=|S_v^t|=L$.

\subsection{Encoder Layer}
In this section, we exposition details about how to convert the extracted interactions from temporal graph learning to time series analysis. These interactions, construed as temporally continuous events, embody four encodings: node, link, time, and frequency. Together, these constitute the comprehensive representation of $X^t_*$, where $*$ refers to either of the nodes $u$ or $v$. 

\subsubsection{Node/Edge Encoding.}
In the realm of temporal graphs, both vertices (nodes) and interactions (edges or links) often harbor concomitant features. To delineate the embeddings allied with interactions, it is imperative to harvest the inherent traits of proximate nodes and edges considering the sequence, designated as $S^t_*$.
Aligning our methodology with established approaches in the literature~\cite{rossi2020temporal}, we adopt a schema to encode nodes and links correspondingly as $X^t_{*, N} \in \mathbb{R}^{L \times d_N}$ and $X^t_{*, E} \in \mathbb{R}^{L \times d_E}$, wherein $d_N$ and $d_E$ typify the dimensions associated with the specific embeddings of nodes and edges, respectively. The specific schema is defined as:
\begin{equation}
X^t_{*, N} = \operatorname{MLP}(x^t_{*, N}), 
X^t_{*, E} = \operatorname{MLP}(x^t_{*, E}),
\label{eq: node}
\end{equation}
where $x^t_{*, N}, x^t_{*, E}$ represents the node $*$ interact nodes' features and the interact link features of $S_*^t$. If the original features do not exist, they are all set to zero.

\subsubsection{Time Encoding.} 
Our schema for time-encoding deploys a cosine function, denoted as $\cos(\Delta t\omega)$, where $\omega = \left\{\alpha^{-(i-1)/\beta}\right\}^{d_T}_{i=1}$. This function is tasked with transliterating static timestamps into vectors represented as $X^t_{*, T} \in \mathbb{R}^{L \times d_T}$. Here, the symbol $d_T$ signifies the dimensionality of time embeddings, and $\alpha$ and $\beta$ function as adjustable parameters. Notably, these parameters ensure that the value of $t_{max} \times \alpha^{-(i-1)/\beta}$ converges to zero when $i$ approaches $d_T$. So the time encoding can be defined as:
\begin{equation}
    X^t_{*, T} = [\cos(\Delta t_1\omega), \cos(\Delta t_2\omega), \cdots, \cos(\Delta t_L\omega)]^T.
\label{eq: time}
\end{equation}

In Eq.~\eqref{eq: time}, we adopt relative timestamps for encoding, in preference to absolute ones. In other words, if an interaction occurs at timestamp $t_0$, and we're to predict an interaction at a specific timestamp $t$, the $\Delta t = t - t_0$ is computed. Importantly, $\omega$ is held constant throughout the training phase, thereby accelerating the process of model optimization.
Furthermore, the adoption of relative time encoding is instrumental in discerning repetitive temporal patterns. This function inspects the temporal gaps between interactions, enabling the construction of a similarity index for comparable timestamps,  contributing to the precise temporal differentiation.

\subsubsection{Frequency Encoding.}
Most existing methods often undervalue the potential for re-interactions with historical nodes and overlook inherent node correlations. We address this gap via a paradigm shift: the introduction of a node interaction frequency encoding technique. This novel approach posits that the frequency of a node's interaction within a historical sequence signifies its relevance. Our method goes beyond analyzing merely historical interaction sequences and incorporates the frequency of both interaction node occurrence and node pair interaction. This judicious approach effectively captures correlations between two nodes' common interaction nodes within their respective historical interaction sequences.

As illustrated in Fig.~\ref{fig: model}, we calculate the frequency of each interaction in both $S_u^t$ and $S_v^t$, represented as $F_u^t \in \mathbb{R}^{L \times 2}$ and $F_v^t \in \mathbb{R}^{L \times 2}$ respectively. This is computed as follows, with the first column denoting the number of times the node appears in $S_u^t$ and the second column denoting the number of times the node appears in $S_v^t$. Specifically, as shown in Fig.~\ref{fig: model}, for the interaction nodes $\{u, b, b\}$ and $\{v, d, b\}$ appearing in $S_u^t$ and $S_v^t$, respectively, then $F_u^t = [[1, 2, 2], [0, 1, 1]]$, and $F_v^t = [[0, 0, 2], [1, 1, 1]]$. We then code the frequency of the node pair interaction, deriving node interaction frequency features for $u$ and $v$, denoted by $X^t_{*, F} \in \mathbb{R}^{L \times d_F}$, wherein $d_F$ means the dimensions of frequency embedding, respectively. This encoding manifests mathematically as follows:
\begin{equation}
    X^t_{*, F} = f(F^t_*[:, 0]) + f(F^t_*[:, 1]).
\label{eq: frequency}
\end{equation}
We implement a two-layer $\text{MLP}$ function with $\text{ReLU}$ activation, designated as $f(\cdot)$, to encode the nodes' frequency interaction features. 

The frequency Encoding fortifies our hypothesis that nodes with a high frequency of shared historical interaction nodes are likelier to interact in the future. Concurrently, they accommodate the variability of recurrence patterns apparent across different network domains and structures.

\subsubsection{Transition to Time Series Analysis.}
Transitioning next to the realm of time series analysis, we employ the encodings formulated earlier. All these encodings are amalgamated and then projected into a trainable weight space $W_\diamond \in \mathbb{R}^{d_\diamond \times d}$ along with $b_\diamond \in \mathbb{R}^{d}$ to yield embeddings $X^t_{u,\diamond} \in \mathbb{R}^{L \times d}$ and $X^t_{v,\diamond} \in \mathbb{R}^{L \times d}$. It's imperative to note that the symbol $\diamond$ can be a representative of $N, E, T$ or $F$. These alignments can be mathematically explained as:
\begin{equation}
\begin{aligned}
    X^t_{u, \diamond} &= X^t_{u, \diamond}W_\diamond + b_\diamond \in \mathbb{R}^{L \times d},\\
    X^t_{v, \diamond} &= X^t_{v, \diamond}W_\diamond + b_\diamond \in \mathbb{R}^{L \times d}.
\end{aligned}
\label{eq: align}
\end{equation}
We then focus on the concatenation of these encoded nodes which can be calculated as follows:
\begin{equation}
\begin{aligned}
X^t_u &= X^t_{u, N} || X^t_{u, E} || X^t_{u, T} || X^t_{u, F} \in \mathbb{R}^{L \times 4d},\\
X^t_v &= X^t_{v, N} || X^t_{v, E} || X^t_{v, T} || X^t_{v, F} \in \mathbb{R}^{L \times 4d}.
\end{aligned}
\label{eq: connect}
\end{equation}
Following this, we treat $X^t_u$ and $X^t_v$ as a time series of length $L$, thus positioning us for a comprehensive analysis of the time series.

\begin{figure}[!ht]
    \centering
    \includegraphics[width=0.75\linewidth]{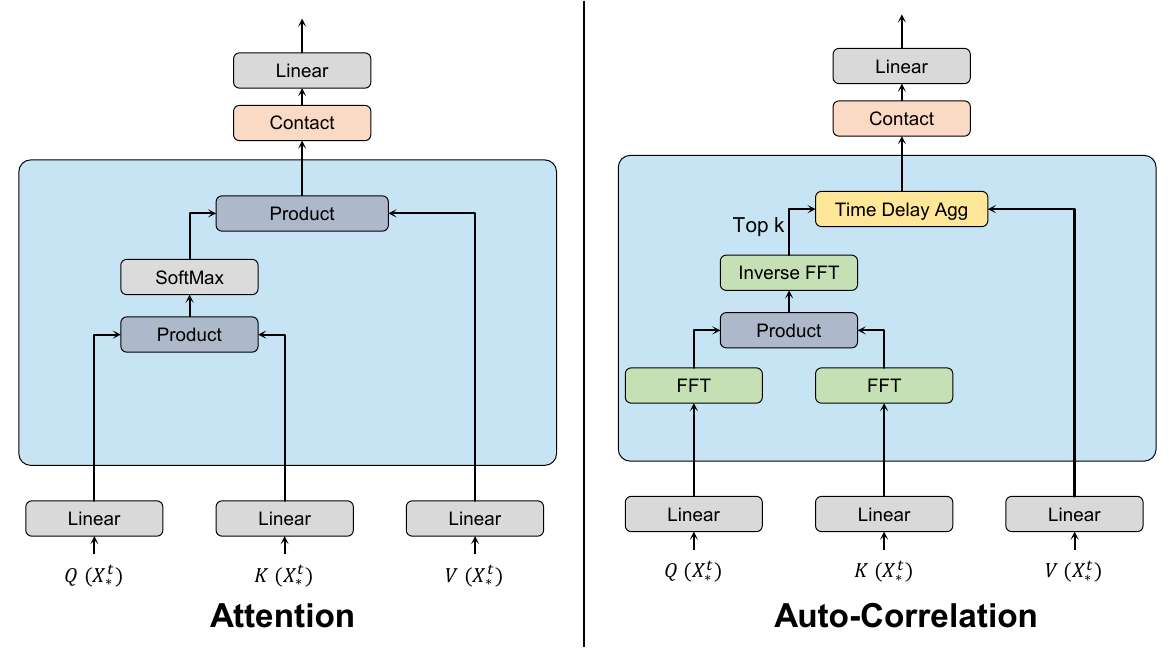}
    \caption{ 
    Attention (left) and ACoM (right). We utilize the Fast Fourier Transform (FFT) to calculate the ACoM, which reflects the time-delay similarities.}
    \label{fig: attention}
\end{figure}
\vspace{-0.12in}

\begin{algorithm*}[!ht]
  \caption{Auto-Correlation Mechanism}
  \label{alg: acom}
  \begin{algorithmic}[1]
  \renewcommand{\algorithmicrequire}{\textbf{Input:}}
    \renewcommand{\algorithmicensure}{\textbf{Output:}}
    \REQUIRE Queries $Q$, Keys $K$, Values $V \in \mathbb{R}^{L\times 4d}$; Number of heads $h$; Hyper-parameter $c$.
    \STATE $Q, K, V \leftarrow \text{Reshape}(Q), \text{Reshape}(K), \text{Reshape}(V)$ 
    \STATE $\mathcal{Q}=\text{FFT}(Q,\text{dim=0}),\mathcal{K}=\text{FFT}(K,\text{dim=0}) $
    \STATE $\text{Corr}=\text{IFFT}\Big(\mathcal{Q}\times\text{Conj}(\mathcal{K}),\text{dim=0}\Big)$
    \STATE $\text{W}_{\text{topk}},\text{I}_{\text{topk}}=\text{Topk}(\text{Corr},\lfloor c\times{\log L}\rfloor, \text{dim=0})$
    \STATE $\text{W}_{\text{topk}}=\text{Softmax}(\text{W}_{\text{topk}}, \text{dim=0})$
    \STATE ${\text{Index}}=\text{Repeat}\Big(\text{arange}(L)\Big)$
    \STATE $V=\text{Repeat}(V)$
    \STATE $\mathcal{R}=\Big[{\text{W}_{\text{topk}}}_{i,:,:}\times\text{gather}\Big(V, ({\text{I}_{\text{topk}}}_{i,:,:}+\text{Index})\Big)\ \textbf{for}\ i\ \textbf{in} \ \text{range}(\lfloor c\times{\log L}\rfloor)\Big]$
    \STATE $\mathcal{R}=\text{Sum}\Big(\text{Stack}(\mathcal{R},\text{dim=0}), \text{dim=0}\Big)$
    \STATE $\textbf{Return}\ \mathcal{R}$ 
  \end{algorithmic} 
\end{algorithm*} 

\subsection{Series Transformer Layer}

In this section, we will describe how to use the Series Transformer layer, including auto-correlated mechanism and Feed-forward networks.

\subsubsection{Auto-Correlation Mechanism.}
As depicted in Fig.~\ref{fig: attention}, we propose an ACoM that utilizes node interaction sequences with series-wise connections to optimize information usage. The ACoM uncovers period-based dependencies by calculating the auto-correlation of series and amalgamating similar sub-sequences through time delay aggregation. We briefly describe the procedure of ACoM in Algorithm~\ref{alg: acom}.

\begin{figure}[!ht]
    \centering
    \includegraphics[width=0.75\linewidth]{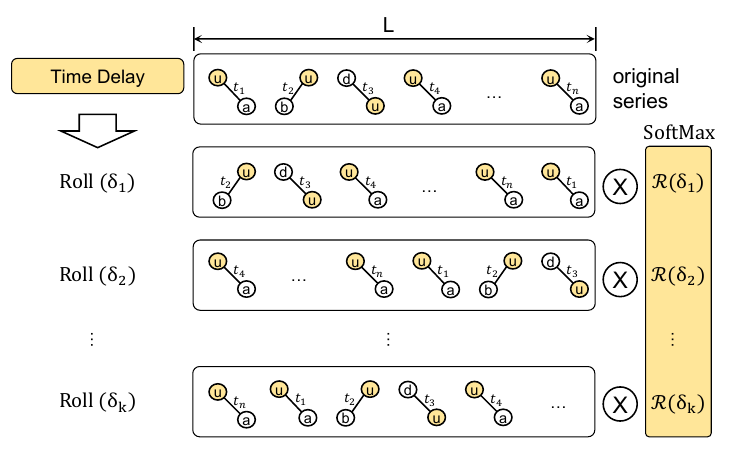}
    \caption{
    Time-delay aggregation block. $\mathcal{R}(\delta)$ reflects the time-delay similarities. Then the similar sub-processes are rolled to the same index based on selected delay $\delta$ and aggregated by $\mathcal{R}(\delta)$.}
    \label{fig: timeDelay}
\end{figure}

\textbf{Period-based Dependencies.}
Inspired by the theory of stochastic processes, we observe that identical phase positions across periods naturally dictate similar sub-processes. Thus, we treat each interaction sequence as a discrete-time process $\left\{\mathcal{X}^t_i\right\}_{i=1}^L$, where $\mathcal{X}^t_i$ is the $i$-th row of $X^t_*$, i.e., the feature of the $i$-th sampled interaction. Then its auto-correlation, $ \mathcal{R}_{\mathcal{X},\mathcal{X}}(\delta)$, can be computed as follows:
\begin{equation}
    \mathcal{R}_{\mathcal{X},\mathcal{X}}(\delta)=\lim_{L\to\infty}\frac1L\sum_{i=1}^{L}\mathcal{X}^t_i\mathcal{X}^{t}_{i-\delta}.
\label{eq: Rxx}
\end{equation}
In this equation, $ \mathcal{R}_{\mathcal{X},\mathcal{X}}(\delta)$ mirrors the time-delay similarity between ${\mathcal{X}^t_i}$ and the lag series, ${\mathcal{X}^t_{i-\delta}}$. As shown in Fig.~\ref{fig: timeDelay}, we employ the auto-correlation $\mathcal{R}(\delta)$ as the unnormalized confidence of the estimated period length $\delta$. We then select the most probable $k$ period lengths, $\delta_1,\cdots, \delta_k$. The resulting period-based dependencies, drawn from the aforementioned estimated periods, can be weighted by the corresponding auto-correlation.

\textbf{Time Delay Aggregation.}
Building on these period-based dependencies, it follows that these dependencies interconnect the sub-sequences across projected periods. In response, we introduce the time delay aggregation block (illustrated in Fig.~\ref{fig: timeDelay}) that allows for the series to roll based on the selected time delay, $\delta_1,\cdots, \delta_k$. This operation aligns similar sub-series onto the same phase position of estimated periods, a technique that contrasts with the point-wise dot-product aggregation method used by the attention family. Lastly, we complete the process by incorporating the softmax-normalized confidences to aggregate the sub-sequences.

We initiate our discourse with a single-head scenario involving a time series $\mathcal{X}^t_i$ of length-$L$. To analogize attention, we use symbols $Q= K= V = X_*^t$, which are derived from the encoder layer. Thus, it can seamlessly interchange the attention mechanism. 

The subsequent step involves introducing the ACoM. This mechanism incorporates several components, with the initial process given by:

\begin{equation}
\delta_{1},\cdots,\delta_{k} =\underset{\delta\in\{1,\cdots,L\}}{\operatorname*{\arg\text{Topk}} }\left(\mathcal{R}_{{Q},{K}}(\delta)\right),
\label{eq: topk}
\end{equation}
where $\operatorname{arg\text{Topk}}(\cdot)$ retrieves the arguments of $\text{Topk}$ auto-correlations, and $k=\left\lfloor c\times\log L\right\rfloor$. Here, $c$ represents a hyper-parameter. 
We then normalize these auto-correlations using a SoftMax function, as shown in the equation:


\begin{equation}
\begin{aligned}
    &\widehat{\mathcal{R}}_{Q,K}(\delta_1),\cdots,\widehat{\mathcal{R}}_{Q,K}(\delta_k) 
    = \text{SoftMax}\left(\mathcal{R}_{Q,K}(\delta_1),\cdots,\mathcal{R}_{Q,K}(\delta_k)\right).
\end{aligned}
\end{equation}
Finally, the ACoM culminates with the equation:


\begin{equation}
\begin{aligned}
\operatorname{ACoM}({Q},{K},{V}) 
=\sum_{i=1}^k\operatorname{Roll}({V},\delta_i)\widehat{\mathcal{R}}_{{Q},{K}}(\delta_i),
\end{aligned}
\label{eq: AC}
\end{equation}
where $\operatorname{Roll}(\mathcal{X}^t,\delta)$ signifies an operation on $\mathcal{X}^t$ with time delay $\delta$. This operation reinstates any elements shifted beyond the initial position at the end of the sequence.

Advancing to the multi-head version employed in auto-correlation, we handle hidden variables with $4d$ channels and $h$ heads. For each $i$-th head, the query, key, and value are designated as $Q_i, K_i, V_i \in R^{L \times \frac{4d}{h}}, i \in \{1, \cdots, h\}$. The operation progression results into:

\begin{equation}
\begin{aligned}
O &= \operatorname{MultiHead}({Q},{K},{V}) 
=W_O\text{Concat}(O_1, O_2, \cdots,O_h),
\end{aligned}
\label{eq: MAC}
\end{equation}
where $O_i =\text{ACoM}({Q}_i,{K}_i,{V}_i)$ from Eq.~\eqref{eq: AC}, $W_O \in \mathbb{R}^d$ is a trainable parameter.


\textbf{Efficient Computation.} For period-based dependencies are inherently sparse, linking to sub-processes at identical phase positions within specific periods, prioritizing most probable delays and hence preventing the selection of converse phases. This mechanism, exhibiting computational aptness, aggregates series equivalent to $O(\log L)$, each measuring $L$ in length. The complexity of Eq.~\eqref{eq: AC} and Eq.~\eqref{eq: MAC} is $O(L\log L)$, illustrating computational efficiency.
\label{sec: efficient}

Auto-correlation calculation (Eq.~\eqref{eq: Rxx}) hinges on Fast Fourier Transforms (FFT), according to the Wiener–Khinchin theorem~\cite{duhamel1990fast}, for a time series $\left\{\mathcal{X}^t_i\right\}_{i=1}^L$. The resultant $\mathcal{R}_{\mathcal{X},\mathcal{X}}(\delta)$ can be deciphered via the following integral equations:


\begin{equation}
\begin{aligned}
    \mathcal{S}_{\mathcal{X},\mathcal{X}}(f) &=\mathcal{F}\left(\mathcal{X}_i^t\right)\mathcal{F}^*\left(\mathcal{X}^t_i\right) = \int_{-\infty}^\infty\mathcal{X}_i^te^{-2\pi if}\mathrm{d}i\overline{\int_{-\infty}^\infty\mathcal{X}_i^te^{-2\pi if}\mathrm{d}i}, \\
\end{aligned}
\label{eq: FFT}
\end{equation}
\begin{equation}
\begin{aligned}
    \mathcal{R}_{\mathcal{X},\mathcal{X}}(\delta) &= \mathcal{F}^{-1}\left(\mathcal{S}_{\mathcal{X},\mathcal{X}}(f)\right)  = \int_{-\infty}^\infty\mathcal{S}_{\mathcal{X},\mathcal{X}}(f)e^{2\pi f\delta}\mathrm{d}f,
\end{aligned}
\label{eq: IFFT}
\end{equation}
where $\delta$ belongs to the set $\left\{1, \cdots, L\right\}$, $\mathcal{F}$ signifies FFT, its inverse is $\mathcal{F}^{-1}$, and $*$ denotes the conjugate operation. By using FFT, the auto-correlation of all lags in $\left\{1, \cdots, L\right\}$ can be computed concurrently, enhancing computational efficiency with complexity at $O(L\log L)$.


\subsubsection{Feed-Forward Network.}

Our Series Transformer layer emanates from the paradigm of the conventional Transformer encoder, as illustrated in~\cite{vaswani2017attention}. A unique adaptation within our implementation involves the placement of the layer normalization (LN) before, rather than after~\cite{xiong2020layer}, the multi-head ACoM, and the feed-forward blocks (FFN). This strategic alteration, now prevalent amongst contemporary Transformer implementations, ensures more effective optimization, as advocated by~\cite{narang2021transformer}.

Particularly in the context of the FFN sub-layer, we maintain a uniform dimensionality of the input, output, and inner layers, matching the dimensional structure with $4d$. The operational functionality of the Series Transformer layer can thus be formally expounded as follows:
\begin{equation}
Z'^{(j)}_* =\operatorname{MultiHead}({Q},{K},{V})+Z^{(j-1)}_*,
\end{equation}
\begin{equation}
Z^{(j)}_* =\operatorname{FFN}(\operatorname{LN}(Z'^{(j)}_*))+Z'^{(j)}_*.
\end{equation}
where ${Q},{K},{V}$ all equal $Z^{(j-1)}_*$. The input of the first layer is $Z_*^{0} = X_*^t \in \mathbb{R}^{L \times 4d}$, and the output of the $J$-th layer is denoted by $H_*^t = Z_*^{(J)} \in \mathbb{R}^{L \times 4d}$. $1 \le j \le J$ denotes the transformer layer number.

\subsection{Adaptive Readout.}
For the output matrix $H_*^t \in \mathbb{R}^{L \times 4d}$ of a node, $H^t_{*,1} \in \mathbb{R}^{1 \times 4d}$ is the token representation of the node interacting with itself and $H^t_{*,l} \in \mathbb{R}^{1 \times 4d}$ is its $l$-th event representation. We calculate the normalized attention coefficients for its $l$-th event: 
\begin{equation}
\lambda_l = \frac{\exp((H^t_{*,1}||H^t_{*,l})W^T_a)}{\sum_{i=2}^{L} \exp((H^t_{*,1}||H^t_{*,i})W^T_a)},
\end{equation}
where $W_a \in \mathbb{R}^{1 \times 8d}$ denotes the learnable projection and $l = 2, \cdots, L$. Therefore, the readout function takes the correlation between each event and the node representation into account. The node representation is finally aggregated as follows:
\begin{equation}
h_*^t = H^t_{*,1} + \sum_{l=2}^{L} \lambda_l H^t_{*,l}.
\label{eq: readout}
\end{equation}

\subsection{Loss Function}
For link prediction loss, we adopt binary cross-entropy loss function, which is defined as:
\begin{equation}
    \mathcal{L}_p = -\sum_{i=1}^S(y_i \log\hat{y}_i+(1-y_i)\log(1-\hat{y}_i)),
    \label{eq: loss}
\end{equation}
\begin{equation}
\begin{aligned}
\hat{y}=\operatorname{Softmax}(\operatorname{MLP}(\operatorname{RELU}(\operatorname{MLP}(h_u^t||h_\upsilon^t)))),
\label{eq: predict}
\end{aligned}
\end{equation}
where $y_i$ represents the ground-truth label of $i$-th sample and the $\hat{y}_i$ uses two nodes' representation represents the prediction value. 
\begin{algorithm}[!ht]

    \renewcommand{\algorithmicrequire}{\textbf{Input:}}
    \renewcommand{\algorithmicensure}{\textbf{Output:}}
    \caption{Training pipeline for TGFormer}
    \begin{algorithmic}[1] %
    \label{alg: all}
        \REQUIRE  A temporal graph $\mathcal{G}$, a node pair $(u, v)$ with a specific timestamp $(t)$, the neighbor sample number $L$, maximum training epoch of 200, early stopping strategy with patience = 20. 
	\ENSURE The probability of the node pair interacting at timestamp $t$. 
        \STATE initial patience = 0;
        \FOR{training epoch = 1, 2, 3, $\dots$}
            \STATE Acquire the $L$ most recent first-hop interaction neighbors of nodes $u$ and $v$ from $\mathcal{G}$ prior to timestamp $t$ as $S^t_u $ and $S^t_v$;
            \FOR{$S^t_u $ and $S^t_v$ in parallel}
                \STATE Obtain node encoding $X^t_{*,N}$ and edge encoding $X^t_{*,E}$ from Eq.~\eqref{eq: node};
                \STATE Obtain time encoding $X^t_{*,T}$ from Eq.~\eqref{eq: time};
                \STATE Obtain frequency encoding $X^t_{*,F}$ from Eq.~\eqref{eq: frequency};
                \STATE $X^t_* \leftarrow X^t_{*,N} || X^t_{*,E} || X^t_{*,T} || X^t_{*,F}$;
                \STATE Initial $Z_*^{(0)} \leftarrow X^t_*$;
                \FOR{Series Transformer Layer $j$}
                    \STATE $Q, K, V \leftarrow X^t_*$;
                    \STATE $Z'^{(j)}_* \leftarrow ACoM(Q, K, V) + Z^{(j-1)}_*$;
                    \STATE $Z^{(j)}_* \leftarrow FFN(LN(Z'^{(j)}_* + Z'^{(j-1)}_*))$;
                \ENDFOR
                \STATE Adaptive readout with Eq.~\eqref{eq: readout};
            \ENDFOR
            \STATE Conduct link prediction with Eq.~\eqref{eq: predict};
            \STATE Compute loss $\mathcal{L}_p$ with Eq.~\eqref{eq: loss};
            \IF{current epoch's metrics worse than the previous epoch's}
                \STATE patience = patience + 1
            \ELSE
                \STATE Save the model parameters from the current epoch;
            \ENDIF
            \IF{patience = 20}
                \STATE Exit training process;
            \ENDIF
        \ENDFOR
     \end{algorithmic}
\end{algorithm}

\subsection{Complexity Analysis}

To analyze the time complexity of TGFormer, we first present its pseudo-code as shown in Algorithm~\ref{alg: all}. 
For simplicity, the representation dimensions for both input and hidden features are denoted by $d$, and $M$ signifies the size of temporal edges $|\mathcal{E}|$. The sampling process responsible for acquiring the $L$ most recent neighbors achieves a time complexity of $O(1)$. Modules can be executed in parallel within the encoder layer, where the encoder complexity reaches a maximum of $O(dL)$. Furthermore, the series transformer layer demands significant computational resources; however, we mitigate the time complexity through the use of subsequences (as mentioned in Section~\ref{sec: efficient}), reducing it to $O(dL\log L)$. Consequently, the overall time complexity of TGFormer is at most $O(MdL\log L)$.

\begin{table*}[!ht]
\caption{Statistics of the datasets.}
\centering
  \resizebox{\linewidth}{!}
  {%
  \begin{tabular}{c|ccccccc}
\hline
Datasets    & Domains     & \#Nodes & \#Links   & \#Node \& Link Features & Bipartite & Time Granularity &Duration      \\ \hline
Wikipedia   & Social      & 9,227  & 157,474   & -- \& 172                & True    & Unix timestamp & 1 month       \\
Reddit      & Social      & 10,984 & 672,447   & -- \& 172                & True    & Unix timestamp  & 1 month       \\
LastFM      & Interaction & 1,980  & 1,293,103 & -- \& --                 & True    & Unix timestamp  & 1 month       \\
Enron       & Social      & 184    & 125,235   & -- \& --                 & False   & Unix timestamp  & 3 years       \\
UCI         & Social      & 1,899  & 59,835    & -- \& --                 & False   & Unix timestamp  & 196 days      \\
CollegeMsg & Social & 1,899  & 59,835 & -- \& -- & False   & Unix timestamp  & 193 days      \\ \hline
\end{tabular}
  }
\label{tab: datasets}
\end{table*}
\section{Experiments}


\subsection{Datasets}
We evaluate the performance of TGFormer on temporal link prediction tasks utilizing six publicly available temporal graph datasets: Wikipedia, Reddit, LastFM, Enron, UCI, and CollegeMsg. The statistical characteristics of these datasets are comprehensively detailed in Table~\ref{tab: datasets}. Further details of the datasets are shown in~\ref{sec: dataset}.

\subsection{Baselines}
Our model is compared with ten state-of-the-art methods on temporal graphs. They are based on sequential models (JODIE~\cite{kumar2019predicting}, DyRep~\cite{trivedi2019dyrep}), memory networks (TGN~\cite{rossi2020temporal}), random walks(CAWN~\cite{wang2021inductive}), attention mechanisms(TGAT~\cite{xu2020inductive}, TCL~\cite{wang2021tcl}, DyGFormer~\cite{yu2023towards}), and others(EdgeBank~\cite{poursafaei2022towards}, GraphMixer~\cite{cong2023we}, FreeDyG~\cite{tian2024freedyg}). Further details of baselines are shown in~\ref{sec: baseline}.

\subsection{Evaluation Tasks and Metrics}

We closely follow~\cite{yu2023towards} by evaluating the model performance for temporal link prediction, which entails forecasting the probability of a link formation between two specified nodes at a given timestamp. This evaluation is conducted under two distinct settings: the transductive setting, which aims to predict future links among nodes observed during the training phase, and the inductive setting, which endeavors to predict future links involving previously unseen nodes. To this end, a multi-layer perceptron is utilized, which takes the concatenated representations of the two nodes as input and outputs the likelihood of a link. Evaluation metrics employed include Average Precision (AP) and Area Under the Receiver Operating Characteristic Curve (AUC-ROC). For the node classification task, AUC-ROC is adopted as the performance metric. All results presented are the aggregate of 10 independent experimental runs.

To provide a more comprehensive comparison, we employ three negative sampling strategies for evaluating temporal link prediction, namely, random, historical, and inductive negative sampling, as proposed in~\cite{poursafaei2022towards}. These strategies differ in the manner in which negative edges are selected, with the latter two imposing more stringent constraints. Specifically, random negative sampling selects negative edges by randomly pairing nodes from the entire graph, ensuring broad coverage of possible negative interactions. In contrast, historical negative sampling restricts negative edges to those that have previously appeared in past timestamps but are absent in the current step, thereby assessing a model's ability to predict edge reoccurrence patterns. Lastly, inductive negative sampling samples negative edges exclusively from previously unseen edges during training, focusing on the model's capacity to generalize to novel structures. For all tasks, datasets are chronologically split into 70\%, 15\%, and 15\% for training, validation, and testing, respectively. Hyperparameter configurations for baseline models strictly follow those outlined in their respective publications, consistent with~\cite{yu2023towards}.

\subsection{Implementatoin Details}
For both tasks, all models undergo a training regimen of up to 100 epochs, incorporating the early stopping strategy with a patience parameter set to 10. The model achieving optimal performance on the validation set is selected for subsequent testing. The Adam optimizer is uniformly employed across all models, and uses supervised binary cross-entropy loss as the objective function, maintaining a consistent learning rate of 0.00001 and a batch size of 200. We conduct our experiments on a machine with the Intel(R) Xeon(R) Gold 6330 CPU @ 2.00GHz with 256 GiB RAM, and four NVIDIA 3090 GPU cards, which is implemented in Python 3.9 with Pytorch. See our code for more details are in https://github.com/isfs/TGFormer.

\begin{table*}[!ht]
    \centering
    \caption{AP(\%) for \emph{transductive} temporal link prediction with random, historical, and inductive negative sampling strategies. NSS is the abbreviation of Negative Sampling Strategies.}
   \resizebox{0.95\linewidth}{!}{
      \begin{tabular}{c|c|ccccccccccc}
      \midrule
      {NSS} & {Datasets} & JODIE        & DyRep        & TGAT         & TGN                   & CAWN   & EdgeBank   & TCL          & GraphMixer   & DyGFormer & FreeDyG  & TGFormer \\
    \midrule
    \multirow{7}*{\rotatebox{90}{Random}}
& {Wikipedia}  & {96.50 ± 0.14} & {94.86 ± 0.06} & {96.94 ± 0.06} & {98.45 ± 0.06} & {98.76 ± 0.03} & {90.37 ± 0.00}          & {96.47 ± 0.16} & {97.25 ± 0.03} & {99.03 ± 0.02}       & \underline{99.26 ± 0.01}    & {\textbf{99.79 ± 0.10}} \\
    & {Reddit}     & {98.31 ± 0.14} & {98.22 ± 0.04} & {98.52 ± 0.02} & {98.63 ± 0.06} & {99.11 ± 0.01} & {94.86 ± 0.00}          & {97.53 ± 0.02} & {97.31 ± 0.01} & {99.22 ± 0.01}       & \underline{99.48 ± 0.01}    & {\textbf{99.86 ± 0.01}} \\
    & {LastFM}     & {70.85 ± 2.13} & {71.92 ± 2.21} & {73.42 ± 0.21} & {77.07 ± 3.97} & {86.99 ± 0.06} & {79.29 ± 0.00}          & {67.27 ± 2.16} & {75.61 ± 0.24} & {\underline{93.00 ± 0.12}} & 92.15 ± 0.16          & {\textbf{96.87 ± 0.34}} \\
    & {Enron}      & {84.77 ± 0.30} & {82.38 ± 3.36} & {71.12 ± 0.97} & {86.53 ± 1.11} & {89.56 ± 0.09} & {83.53 ± 0.00}          & {79.70 ± 0.71} & {82.25 ± 0.16} & {\underline{92.47 ± 0.12}} & 92.51 ± 0.05          & {\textbf{97.14 ± 0.72}} \\
    & {UCI}        & {89.43 ± 1.09} & {65.14 ± 2.30} & {79.63 ± 0.70} & {92.34 ± 1.04} & {95.18 ± 0.06} & {76.20 ± 0.00}          & {89.57 ± 1.63} & {93.25 ± 0.57} & {95.79 ± 0.17}       & \underline{96.28 ± 0.11}    & {\textbf{99.09 ± 0.17}} \\
    & {CollegeMsg} & {75.41 ± 1.82} & {56.92 ± 6.03} & {80.27 ± 0.29} & {92.62 ± 0.99} & {95.86 ± 0.06} & {76.42 ± 0.00}          & {83.64 ± 0.10} & {93.04 ± 0.29} & {95.79 ± 0.02}       & \underline{95.81 ± 0.22}    & {\textbf{99.17 ± 0.22}} \\ \cmidrule{2-13}
    & {Avg. Rank}  & 8.17 & 9.5  & 8.17 & 5.5  & 3.67 & 8.83          & 8.83 & 7    & 3          & \underline{2.33}            & \textbf{1}    \\ \midrule
    \multirow{7}*{\rotatebox{90}{Historical}}
  & {Wikipedia}  & {83.01 ± 0.66} & {79.93 ± 0.56} & {87.38 ± 0.22} & {86.86 ± 0.33} & {71.21 ± 1.67} & {73.35 ± 0.00}          & {89.05 ± 0.39} & {90.90 ± 0.10} & {82.23 ± 2.54}       & \underline{91.59 ± 0.57}    & {\textbf{92.47 ± 0.19}} \\
    & {Reddit}     & {80.03 ± 0.36} & {79.83 ± 0.31} & {79.55 ± 0.20} & {81.22 ± 0.61} & {80.82 ± 0.45} & {73.59 ± 0.00}          & {77.14 ± 0.16} & {78.44 ± 0.18} & {81.57 ± 0.67}       & \textbf{85.67 ± 1.01} & {\underline{84.44 ± 1.49}}    \\
    & {LastFM}     & {74.35 ± 3.81} & {74.92 ± 2.46} & {71.59 ± 0.24} & {76.87 ± 4.64} & {69.86 ± 0.43} & {73.03 ± 0.00}          & {59.30 ± 2.31} & {72.47 ± 0.49} & {\underline{81.57 ± 0.48}} & 79.71 ± 0.51          & {\textbf{84.52 ± 0.67}} \\
    & {Enron}      & {69.85 ± 2.70} & {71.19 ± 2.76} & {64.07 ± 1.05} & {73.91 ± 1.76} & {64.73 ± 0.36} & {76.53 ± 0.00}          & {70.66 ± 0.39} & {77.98 ± 0.92} & {75.63 ± 0.73}       & \underline{78.87 ± 0.82}    & {\textbf{82.64 ± 1.39}} \\
    & {UCI}        & {75.24 ± 5.80} & {55.10 ± 3.14} & {68.27 ± 1.37} & {80.43 ± 2.12} & {65.30 ± 0.43} & {65.50 ± 0.00}          & {80.25 ± 2.74} & {84.11 ± 1.35} & {82.17 ± 0.82}       & \underline{86.10 ± 1.19}    & {\textbf{86.12 ± 1.50}} \\
    & {CollegeMsg} & {64.41 ± 6.52} & {47.73 ± 0.97} & {68.18 ± 1.04} & {80.65 ± 1.13} & {84.54 ± 0.11} & {44.16 ± 0.00}          & {68.53 ± 0.08} & {83.93 ± 0.09} & {80.93 ± 0.37}       & \underline{89.11 ± 0.34}    & {\textbf{90.02 ± 1.19}} \\ \cmidrule{2-13}
    & {Avg. Rank}  & 7.33 & 8.17 & 8.17 & 5.17 & 8.17 & 8.67          & 7.67 & 5    & 4.5        & \underline{2}               & \textbf{1.17} \\ \midrule
    \multirow{7}*{\rotatebox{90}{Inductive}}
    & {Wikipedia}  & {75.65 ± 0.79} & {70.21 ± 1.58} & {87.00 ± 0.16} & {85.62 ± 0.44} & {74.06 ± 2.62} & {80.63 ± 0.00}          & {86.76 ± 0.72} & {88.59 ± 0.17} & {78.29 ± 5.38}       & \underline{90.05 ± 0.79}    & {\textbf{93.10 ± 0.71}} \\
    & {Reddit}     & {86.98 ± 0.16} & {86.30 ± 0.26} & {89.59 ± 0.24} & {88.10 ± 0.24} & {91.67 ± 0.24} & {85.48 ± 0.00}          & {87.45 ± 0.29} & {85.26 ± 0.11} & {\underline{91.11 ± 0.40}} & 90.74 ± 0.17          & {\textbf{92.37 ± 1.64}} \\
    & {LastFM}     & {62.67 ± 4.49} & {64.41 ± 2.70} & {71.13 ± 0.17} & {65.95 ± 5.98} & {67.48 ± 0.77} & {\textbf{75.49 ± 0.00}} & {58.21 ± 0.89} & {68.12 ± 0.33} & {73.97 ± 0.50}       & 72.19 ± 0.24          & {\underline{74.01 ± 1.35}}    \\
    & {Enron}      & {68.96 ± 0.98} & {67.79 ± 1.53} & {63.94 ± 1.36} & {70.89 ± 2.72} & {75.15 ± 0.58} & {73.89 ± 0.00}          & {71.29 ± 0.32} & {75.01 ± 0.79} & {77.41 ± 0.89}       & 77.81 ± 0.65          & {\textbf{84.34 ± 1.74}} \\
    & {UCI}        & {65.99 ± 1.40} & {54.79 ± 1.76} & {68.67 ± 0.84} & {70.94 ± 0.71} & {64.61 ± 0.48} & {57.43 ± 0.00}          & {76.01 ± 1.11} & {80.10 ± 0.51} & {72.25 ± 1.71}       & \textbf{82.35 ± 0.73} & {\underline{80.70 ± 1.16}}    \\
    & {CollegeMsg} & {53.49 ± 0.53} & {54.13 ± 1.64} & {68.54 ± 0.85} & {71.90 ± 0.06} & {75.12 ± 0.09} & {43.49 ± 0.00}          & {68.70 ± 0.02} & {79.17 ± 0.09} & {70.44 ± 0.83}       & \textbf{81.42 ± 1.15} & {\underline{80.15 ± 1.72}}    \\ \cmidrule{2-13}
    & {Avg. Rank}  & 9    & 9.83 & 6.67 & 6.5  & 6    & 7.5           & 6.83 & 5.17 & 4.67       & \underline{2.33}            & \textbf{1.5} \\ \midrule
\end{tabular}}
\label{tab: tap}
\end{table*}

\begin{table*}[!ht]
    \centering
    \caption{AUC-ROC(\%) for \emph{transductive} temporal link prediction with random, historical, and inductive negative sampling strategies. NSS is the abbreviation of Negative Sampling Strategies.}
   \resizebox{0.95\linewidth}{!}{
      \begin{tabular}{c|c|ccccccccccc}
      \midrule
      {NSS} & {Datasets} & JODIE        & DyRep        & TGAT         & TGN                   & CAWN   & EdgeBank   & TCL          & GraphMixer   & DyGFormer  & FreeDyG        & TGFormer \\
    \midrule
    \multirow{7}*{\rotatebox{90}{Random}}
& {Wikipedia}  & {96.33 ± 0.07} & {94.37 ± 0.09} & {96.67 ± 0.07} & {98.37 ± 0.07} & {98.54 ± 0.04} & {90.78 ± 0.00}          & {95.84 ± 0.18} & {96.92 ± 0.03}       & {98.91 ± 0.02}       & \underline{99.41 ± 0.01}    & {\textbf{99.77 ± 0.10}} \\
    & {Reddit}     & {98.31 ± 0.05} & {98.17 ± 0.05} & {98.47 ± 0.02} & {98.60 ± 0.06} & {99.01 ± 0.01} & {95.37 ± 0.00}          & {97.42 ± 0.02} & {97.17 ± 0.02}       & {99.15 ± 0.01}       & \underline{99.50 ± 0.01}    & {\textbf{99.82 ± 0.01}} \\
    & {LastFM}     & {70.49 ± 1.66} & {71.16 ± 1.89} & {71.59 ± 0.18} & {78.47 ± 2.94} & {85.92 ± 0.10} & {83.77 ± 0.00}          & {64.06 ± 1.16} & {73.53 ± 0.12}       & {93.05 ± 0.10}       & \underline{93.42 ± 0.15}    & {\textbf{96.85 ± 0.44}} \\
    & {Enron}      & {87.96 ± 0.52} & {84.89 ± 3.00} & {68.89 ± 1.10} & {88.32 ± 0.99} & {90.45 ± 0.14} & {87.05 ± 0.00}          & {75.74 ± 0.72} & {84.38 ± 0.21}       & {93.33 ± 0.13}       & \underline{94.01 ± 0.11}    & {\textbf{97.13 ± 0.90}} \\
    & {UCI}        & {90.44 ± 0.49} & {68.77 ± 2.34} & {78.53 ± 0.74} & {92.03 ± 1.13} & {93.87 ± 0.08} & {77.30 ± 0.00}          & {87.82 ± 1.36} & {91.81 ± 0.67}       & {94.49 ± 0.26}       & \underline{95.00 ± 0.21}    & {\textbf{98.68 ± 0.30}} \\
    & {CollegeMsg} & {78.87 ± 1.71} & {61.72 ± 6.95} & {79.25 ± 0.76} & {92.39 ± 1.00} & {94.63 ± 0.11} & {77.32 ± 0.00}          & {83.32 ± 0.08} & {91.71 ± 0.28}       & {94.52 ± 0.03}       & \underline{95.07 ± 0.20}    & {\textbf{98.88 ± 0.30}} \\ \cmidrule{2-13}
    & {Avg. Rank}  & 7.83   & 9.5    & 8.17   & 5.17   & 3.83   & 9               & 9      & 7.33         & 3.17         & \underline{2}               & \textbf{1}      \\ \midrule
\multirow{7}*{\rotatebox{90}{Historical}}
& {Wikipedia}  & {80.77 ± 0.73} & {77.74 ± 0.33} & {82.87 ± 0.22} & {82.74 ± 0.32} & {67.84 ± 0.64} & {77.27 ± 0.00}          & {85.76 ± 0.46} & {\underline{87.68 ± 0.17}} & {78.80 ± 1.95}       & 82.78 ± 0.30          & {\textbf{92.07 ± 0.33}} \\
    & {Reddit}     & {80.52 ± 0.32} & {80.15 ± 0.18} & {79.33 ± 0.16} & {81.11 ± 0.19} & {80.27 ± 0.30} & {78.58 ± 0.00}          & {76.49 ± 0.16} & {77.80 ± 0.12}       & {80.54 ± 0.29}       & \textbf{85.92 ± 0.10} & {\underline{83.08 ± 2.43}}    \\
    & {LastFM}     & {75.22 ± 2.36} & {74.65 ± 1.98} & {64.27 ± 0.26} & {77.97 ± 3.04} & {67.88 ± 0.24} & {78.09 ± 0.00}          & {47.24 ± 3.13} & {64.21 ± 0.73}       & {\underline{78.78 ± 0.35}} & 75.74 ± 0.72          & {\textbf{82.39 ± 0.67}} \\
    & {Enron}      & {75.39 ± 2.37} & {74.69 ± 3.55} & {61.85 ± 1.43} & {77.09 ± 2.22} & {65.10 ± 0.34} & {\underline{79.59 ± 0.00}}    & {67.95 ± 0.88} & {75.27 ± 1.14}       & {76.55 ± 0.52}       & 75.74 ± 0.72          & {\textbf{80.59 ± 1.47}} \\
    & {UCI}        & {78.64 ± 3.50} & {57.91 ± 3.12} & {58.89 ± 1.57} & {77.25 ± 2.68} & {57.86 ± 0.15} & {69.56 ± 0.00}          & {72.25 ± 3.46} & {77.54 ± 2.02}       & {76.97 ± 0.24}       & \underline{80.38 ± 0.26}    & {\textbf{83.80 ± 2.67}} \\
    & {CollegeMsg} & {66.92 ± 5.00} & {49.37 ± 1.37} & {58.19 ± 1.11} & {77.65 ± 1.40} & {79.45 ± 0.15} & {34.64 ± 0.00}          & {58.55 ± 0.18} & {77.50 ± 0.16}       & {76.25 ± 0.12}       & \underline{84.83 ± 0.40}    & {\textbf{88.58 ± 1.10}} \\ \cmidrule{2-13}
    & {Avg. Rank}  & 5.67   & 8.5    & 8.33   & 4.33   & 8.17   & 7.33            & 8.17   & 6.33         & 5.17         & \underline{2.67}            & \textbf{1.33}   \\ \midrule
\multirow{7}*{\rotatebox{90}{Inductive}}
& {Wikipedia}  & {70.96 ± 0.78} & {67.36 ± 0.96} & {81.93 ± 0.22} & {80.97 ± 0.31} & {70.95 ± 0.95} & {81.73 ± 0.00}          & {82.19 ± 0.48} & {\underline{84.28 ± 0.30}} & {75.09 ± 3.70}       & 82.74 ± 0.32          & {\textbf{91.01 ± 0.76}} \\
    & {Reddit}     & {83.51 ± 0.15} & {82.90 ± 0.31} & {87.13 ± 0.20} & {84.56 ± 0.24} & {88.04 ± 0.29} & {85.93 ± 0.00}          & {84.67 ± 0.29} & {82.21 ± 0.13}       & {86.23 ± 0.51}       & \underline{84.38 ± 0.21}    & {\textbf{89.42 ± 2.17}} \\
    & {LastFM}     & {61.32 ± 3.49} & {62.15 ± 2.12} & {63.99 ± 0.21} & {65.46 ± 4.27} & {67.92 ± 0.44} & {\textbf{77.37 ± 0.00}} & {46.93 ± 2.59} & {60.22 ± 0.32}       & {69.25 ± 0.36}       & \underline{72.30 ± 0.59}    & {70.54 ± 0.67}          \\
    & {Enron}      & {70.92 ± 1.05} & {68.73 ± 1.34} & {60.45 ± 2.12} & {71.34 ± 2.46} & {75.17 ± 0.50} & {75.00 ± 0.00}          & {67.64 ± 0.86} & {71.53 ± 0.85}       & {74.07 ± 0.64}       & \underline{77.27 ± 0.61}    & {\textbf{81.19 ± 1.21}} \\
    & {UCI}        & {64.14 ± 1.26} & {54.25 ± 2.01} & {60.80 ± 1.01} & {64.11 ± 1.04} & {58.06 ± 0.26} & {58.03 ± 0.00}          & {70.05 ± 1.86} & {74.59 ± 0.74}       & {65.96 ± 1.18}       & \underline{75.39 ± 0.57}    & {\textbf{75.98 ± 2.90}} \\
    & {CollegeMsg} & {53.34 ± 0.22} & {53.13 ± 2.16} & {59.84 ± 1.16} & {65.58 ± 0.42} & {69.28 ± 0.24} & {30.53 ± 0.00}          & {59.67 ± 0.16} & {74.06 ± 0.25}       & {64.77 ± 0.13}       & \underline{76.32 ± 0.67}    & {\textbf{76.92 ± 1.31}} \\ \cmidrule{2-13}
    & {Avg. Rank}  & 8.33   & 9.83   & 6.83   & 6.5    & 5.5    & 6.17            & 7.17   & 5.83         & 5.33         & \underline{3.17}            & \textbf{1.33}  \\ \midrule
\end{tabular}}
\label{tab: tauc}
\end{table*}

\begin{table*}[!ht]
    \centering
    \caption{AP(\%) for \emph{inductive} temporal link prediction with random, historical, and inductive negative sampling strategies. NSS is the abbreviation of Negative Sampling Strategies.}
   \resizebox{0.95\linewidth}{!}{
      \begin{tabular}{c|c|cccccccccc}
      \midrule
      {NSS} & {Datasets} & JODIE        & DyRep        & TGAT         & TGN                   & CAWN      & TCL          & GraphMixer   & DyGFormer & FreeDyG         & \textbf{TGFormer} \\
    \midrule
    \multirow{7}*{\rotatebox{90}{Random}}
& Wikipedia  & 94.82 ± 0.20 & 92.43 ± 0.37 & 96.22 ± 0.07       & 97.83 ± 0.04 & 98.24 ± 0.03       & 96.22 ± 0.17 & 96.65 ± 0.02          & 98.59 ± 0.03       & \underline{98.97 ± 0.01}    & \textbf{99.27 ± 0.25} \\
    & Reddit     & 96.50 ± 0.13 & 96.09 ± 0.11 & 97.09 ± 0.04       & 97.50 ± 0.07 & 98.62 ± 0.01       & 94.09 ± 0.07 & 95.26 ± 0.02          & 98.84 ± 0.02       & \underline{98.91 ± 0.01}    & \textbf{99.78 ± 0.04} \\
    & LastFM     & 81.61 ± 3.82 & 83.02 ± 1.48 & 78.63 ± 0.31       & 81.45 ± 4.29 & 89.42 ± 0.07       & 73.53 ± 1.66 & 82.11 ± 0.42          & 94.23 ± 0.09       & \underline{94.89 ± 0.01}    & \textbf{97.97 ± 0.37} \\
    & Enron      & 80.72 ± 1.39 & 74.55 ± 3.95 & 67.05 ± 1.51       & 77.94 ± 1.02 & 86.35 ± 0.51       & 76.14 ± 0.79 & 75.88 ± 0.48          & \underline{89.76 ± 0.34} & 89.69 ± 0.17          & \textbf{91.88 ± 1.26} \\
    & UCI        & 79.86 ± 1.48 & 57.48 ± 1.87 & 79.54 ± 0.48       & 88.12 ± 2.05 & 92.73 ± 0.06       & 87.36 ± 2.03 & 91.19 ± 0.42          & 94.54 ± 0.12       & \underline{94.85 ± 0.10}    & \textbf{99.16 ± 0.15} \\
    & CollegeMsg & 63.30 ± 1.26 & 52.72 ± 2.58 & 79.58 ± 0.47       & 88.07 ± 1.44 & \underline{94.44 ± 0.05} & 81.18 ± 0.04 & 90.94 ± 0.22          & 94.36 ± 0.12       & 93.99 ± 0.41          & \textbf{97.73 ± 0.22} \\ \cmidrule{2-12}
    & Avg. Rank  & 7.5          & 8.67         & 8.17               & 6            & 3.67               & 8.17         & 6.5                   & 2.83               & \underline{2.5}             & \textbf{1}            \\ \midrule
    \multirow{7}*{\rotatebox{90}{Historical}}
& Wikipedia  & 68.69 ± 0.39 & 62.18 ± 1.27 & \underline{84.17 ± 0.22} & 81.76 ± 0.32 & 67.27 ± 1.63       & 82.20 ± 2.18 & \textbf{87.60 ± 0.30} & 71.42 ± 4.43       & 82.78 ± 0.30          & 78.69 ± 1.25          \\
    & Reddit     & 62.34 ± 0.54 & 61.60 ± 0.72 & 63.47 ± 0.36       & 64.85 ± 0.85 & 63.67 ± 0.41       & 60.83 ± 0.25 & 64.50 ± 0.26          & 65.37 ± 0.60       & \underline{66.02 ± 0.41}    & \textbf{72.95 ± 1.13} \\
    & LastFM     & 70.39 ± 4.31 & 71.45 ± 1.76 & 75.27 ± 0.25       & 66.65 ± 6.11 & 71.33 ± 0.47       & 65.78 ± 0.65 & 76.42 ± 0.22          & 76.35 ± 0.52       & \textbf{77.28 ± 0.21} & \underline{76.81 ± 0.79}    \\
    & Enron      & 65.86 ± 3.71 & 62.08 ± 2.27 & 61.40 ± 1.31       & 62.91 ± 1.16 & 60.70 ± 0.36       & 67.11 ± 0.62 & 72.37 ± 1.37          & 67.07 ± 0.62       & \underline{73.01 ± 0.88}    & \textbf{76.77 ± 1.01} \\
    & UCI        & 63.11 ± 2.27 & 52.47 ± 2.06 & 70.52 ± 0.93       & 70.78 ± 0.78 & 64.54 ± 0.47       & 76.71 ± 1.00 & \underline{81.66 ± 0.49}    & 72.13 ± 1.87       & \textbf{82.35 ± 0.39} & 77.24 ± 1.59          \\
    & CollegeMsg & 50.51 ± 0.75 & 54.43 ± 1.79 & 70.50 ± 1.18       & 71.60 ± 0.31 & 74.14 ± 0.17       & 69.80 ± 0.23 & \underline{80.15 ± 0.18}    & 69.59 ± 1.25       & \textbf{83.31 ± 0.78} & 77.80 ± 1.11          \\ \cmidrule{2-12}
    & Avg. Rank  & 8.17         & 8.67         & 6                  & 6            & 7.33               & 6.5          & 2.67                  & 5.33               & \textbf{1.67}         & \underline{2.67}       \\  \midrule
    \multirow{7}*{\rotatebox{90}{Inductive}}
& Wikipedia  & 68.70 ± 0.39 & 62.19 ± 1.28 & 84.17 ± 0.22       & 81.77 ± 0.32 & 67.24 ± 1.63       & 82.20 ± 2.18 & \textbf{87.60 ± 0.29} & 71.42 ± 4.43       & \underline{87.54 ± 0.26}    & 78.13 ± 1.56          \\
    & Reddit     & 62.32 ± 0.54 & 61.58 ± 0.72 & 63.40 ± 0.36       & 64.84 ± 0.84 & 63.65 ± 0.41       & 60.81 ± 0.26 & 64.49 ± 0.25          & \underline{65.35 ± 0.60} & 64.98 ± 0.20          & \textbf{72.95 ± 1.13} \\
    & LastFM     & 70.39 ± 4.31 & 71.45 ± 1.75 & 76.28 ± 0.25       & 69.46 ± 4.65 & 71.33 ± 0.47       & 65.78 ± 0.65 & \underline{76.42 ± 0.22}    & 76.35 ± 0.52       & \textbf{77.01 ± 0.43} & 74.21 ± 1.02          \\
    & Enron      & 65.86 ± 3.71 & 62.08 ± 2.27 & 61.40 ± 1.30       & 62.90 ± 1.16 & 60.72 ± 0.36       & 67.11 ± 0.62 & 72.37 ± 1.38          & 67.07 ± 0.62       & 72.85 ± 0.81          & \textbf{76.77 ± 1.02} \\
    & UCI        & 63.16 ± 2.27 & 52.47 ± 2.09 & 70.49 ± 0.93       & 70.73 ± 0.79 & 64.54 ± 0.47       & 76.65 ± 0.99 & \underline{81.64 ± 0.49}    & 72.13 ± 1.86       & \textbf{82.06 ± 0.58} & 77.19 ± 1.09          \\
    & CollegeMsg & 50.57 ± 0.76 & 54.47 ± 1.81 & 70.50 ± 1.19       & 71.63 ± 0.31 & 74.11 ± 0.17       & 69.80 ± 0.24 & \underline{80.13 ± 0.18}    & 69.55 ± 1.27       & \textbf{83.33 ± 0.79} & 77.72 ± 1.11          \\  \cmidrule{2-12}
    & Avg. Rank  & 8.17         & 8.67         & 6                  & 6            & 7.33               & 6.5          & \underline{2.5}             & 5                  & \textbf{1.67}         & 3.17      \\   \midrule       
\end{tabular}}
\label{tab: iap}
\end{table*}

\begin{table*}[!ht]
    \centering
    \caption{AUC-ROC(\%) for \emph{inductive} temporal link prediction with random, historical, and inductive negative sampling strategies. NSS is the abbreviation of Negative Sampling Strategies.}
   \resizebox{0.95\linewidth}{!}{
      \begin{tabular}{c|c|cccccccccc}
      \midrule
      {NSS} & {Datasets} & JODIE        & DyRep        & TGAT         & TGN                   & CAWN      & TCL          & GraphMixer   & DyGFormer   & FreeDyG       & TGFormer \\
    \midrule
    \multirow{7}*{\rotatebox{90}{Random}}
& Wikipedia  & 94.33 ± 0.27 & 91.49 ± 0.45 & 95.90 ± 0.09 & 97.72 ± 0.03 & 98.03 ± 0.04       & 95.57 ± 0.20 & 96.30 ± 0.04 & 98.48 ± 0.03 & \underline{99.01 ± 0.02}    & \textbf{99.15 ± 0.28} \\
    & Reddit     & 96.52 ± 0.13 & 96.05 ± 0.12 & 96.98 ± 0.04 & 97.39 ± 0.07 & 98.42 ± 0.02       & 93.80 ± 0.07 & 94.97 ± 0.05 & 98.71 ± 0.01 & \underline{98.84 ± 0.01}    & \textbf{99.69 ± 0.06} \\
    & LastFM     & 81.13 ± 3.39 & 82.24 ± 1.51 & 76.99 ± 0.29 & 82.61 ± 3.15 & 87.82 ± 0.12       & 70.84 ± 0.85 & 80.37 ± 0.18 & 94.08 ± 0.08 & \underline{94.32 ± 0.03}    & \textbf{97.93 ± 0.37} \\
    & Enron      & 81.96 ± 1.34 & 76.34 ± 4.20 & 64.63 ± 1.74 & 78.83 ± 1.11 & 87.02 ± 0.50       & 72.33 ± 0.99 & 76.51 ± 0.71 & \textbf{90.69 ± 0.26} & 89.51 ± 0.20 & \underline{89.92 ± 1.45}    \\
    & UCI        & 78.80 ± 0.94 & 58.08 ± 1.81 & 77.64 ± 0.38 & 86.68 ± 2.29 & 90.40 ± 0.11       & 84.49 ± 1.82 & 89.30 ± 0.57 & 92.63 ± 0.13 & \underline{93.01 ± 0.08}    & \textbf{98.84 ± 0.22} \\
    & CollegeMsg & 66.00 ± 2.05 & 54.09 ± 3.53 & 77.00 ± 0.17 & 86.73 ± 1.82 & 92.29 ± 0.06       & 78.76 ± 0.05 & 88.40 ± 0.18 & 92.21 ± 0.13 & \underline{92.74 ± 0.34}    & \textbf{96.49 ± 0.31} \\ \cmidrule{2-12}
    & Avg. Rank  & 7.5 & 8.67& 8.17& 5.5 & 3.83      & 8.5 & 6.67& 2.83& \underline{2.17}   & \textbf{1.17}\\ \midrule
    \multirow{7}*{\rotatebox{90}{Historical}}
& Wikipedia  & 61.86 ± 0.53 & 57.54 ± 1.09 & 78.38 ± 0.20 & 75.75 ± 0.29 & 62.04 ± 0.65       & 79.79 ± 0.96 & \textbf{82.87 ± 0.21} & 68.33 ± 2.82 & \underline{82.08 ± 0.32}    & 76.75 ± 1.67 \\
    & Reddit     & 61.69 ± 0.39 & 60.45 ± 0.37 & 64.43 ± 0.27 & 64.55 ± 0.50 & 64.94 ± 0.21       & 61.43 ± 0.26 & 64.27 ± 0.13 & 64.81 ± 0.25 & 66.79 ± 0.31 & \textbf{69.26 ± 2.38} \\
    & LastFM     & 68.44 ± 3.26 & 68.79 ± 1.08 & 69.89 ± 0.28 & 66.99 ± 5.62 & 67.69 ± 0.24       & 55.88 ± 1.85 & 70.07 ± 0.20 & \underline{70.73 ± 0.37}    & \textbf{72.63 ± 0.16} & 70.33 ± 1.02 \\
    & Enron      & 65.32 ± 3.57 & 61.50 ± 2.50 & 57.84 ± 2.18 & 62.68 ± 1.09 & 62.25 ± 0.40       & 64.06 ± 1.02 & 68.20 ± 1.62 & 65.78 ± 0.42 & 70.09 ± 0.65 & 68.79 ± 1.58 \\
    & UCI        & 60.24 ± 1.94 & 51.25 ± 2.37 & 62.32 ± 1.18 & 62.69 ± 0.90 & 56.39 ± 0.10       & 70.46 ± 1.94 & \underline{75.98 ± 0.84}    & 65.55 ± 1.01 & \underline{76.01 ± 0.75}    & 67.72 ± 1.97 \\
    & CollegeMsg & 48.57 ± 1.27 & 52.31 ± 1.53 & 61.53 ± 1.45 & 63.89 ± 0.71 & 67.77 ± 0.09       & 60.05 ± 0.08 & \underline{74.54 ± 0.16}    & 63.15 ± 0.44 & \underline{77.55 ± 0.52}    & 74.18 ± 0.49 \\ \cmidrule{2-12}
    & Avg. Rank  & 7.83& 9   & 6.5 & 6.33& 6.67      & 6.5 & 3.17& 4.67& \textbf{1.33}& \underline{3}      \\\midrule
    \multirow{7}*{\rotatebox{90}{Inductive}}
    & Wikipedia  & 61.87 ± 0.53 & 57.54 ± 1.09 & 78.38 ± 0.20 & 75.76 ± 0.29 & 62.02 ± 0.65       & 79.79 ± 0.96 & \underline{82.88 ± 0.21}    & 68.33 ± 2.82 & \textbf{83.17 ± 0.31} & 76.04 ± 2.08 \\
    & Reddit     & 61.69 ± 0.39 & 60.44 ± 0.37 & 64.39 ± 0.27 & 64.55 ± 0.50 & \underline{64.91 ± 0.21} & 61.36 ± 0.26 & 64.27 ± 0.13 & 64.80 ± 0.25 & 64.51 ± 0.19 & \textbf{69.25 ± 2.38} \\
    & LastFM     & 68.44 ± 3.26 & 68.79 ± 1.08 & 69.89 ± 0.28 & 66.99 ± 5.61 & 67.68 ± 0.24       & 55.88 ± 1.85 & 70.07 ± 0.20 & 70.73 ± 0.37 & \underline{71.42 ± 0.33}    & \textbf{73.21 ± 1.42} \\
    & Enron      & 65.32 ± 3.57 & 61.50 ± 2.50 & 57.83 ± 2.18 & 62.68 ± 1.09 & 62.27 ± 0.40       & 64.05 ± 1.02 & 68.19 ± 1.63 & 65.79 ± 0.42 & \underline{68.79 ± 0.91}    & \textbf{72.79 ± 1.59} \\
    & UCI        & 60.27 ± 1.94 & 51.26 ± 2.40 & 62.29 ± 1.17 & 62.66 ± 0.91 & 56.39 ± 0.11       & 70.42 ± 1.93 & \textbf{75.97 ± 0.85} & 65.58 ± 1.00 & \underline{73.41 ± 0.88}    & 72.78 ± 1.34 \\
    & CollegeMsg & 48.64 ± 1.26 & 52.36 ± 1.53 & 61.51 ± 1.47 & 63.93 ± 0.70 & 67.72 ± 0.09       & 60.05 ± 0.10 & \textbf{74.53 ± 0.16} & 63.14 ± 0.44 & \underline{77.58 ± 0.53}    & 73.45 ± 0.51 \\ \cmidrule{2-12}
    & Avg. Rank  & 7.83& 9   & 6.5 & 6.17& 6.5       & 6.67& 3.17& 4.67& \textbf{2.17}& \underline{2.33}  \\ \midrule
\end{tabular}}
\label{tab: iauc}
\end{table*}

\subsection{Link Prediction}
We compare our TGFormer with the previous state-of-the-art in both transductive and inductive link prediction using the AP and AUC-ROC metrics. To provide a more comprehensive study of our TGFormer, we present results for all three negative sampling strategies. 
We put the \emph{transductive} results in Table~\ref{tab: tap} and Table~\ref{tab: tauc} and \emph{inductive} results in Table~\ref{tab: iap} and Table~\ref{tab: iauc}, respectively. The best and second-best results are marked in bold and underlined fonts. Note that EdgeBank~\cite{poursafaei2022towards} can only evaluate transductive temporal link prediction and therefore does not give its results in the inductive setting. From the above table, we can see that TGFormer outperforms the existing methods in most cases, which is far superior to the second one.

We summarize the superiority of TGFormer in two aspects. First, the transformer architecture allows TGFormer to process longer histories and capture long-term temporal dependencies effectively. As illustrated in Fig.~\ref{fig: Long_term}, the input sequence lengths handled by TGFormer are significantly longer than those managed by baseline across most datasets, underscoring TGFormer's superior aptitude in leveraging longer sequences. Second, benefiting from capturing periodic temporal patterns, especially for Wikipedia and Reddit in the case of harder negative sampling, TGFormer also has a small performance degradation compared to the other approaches.

\begin{table}[!htbp]
\centering
\caption{AUC-ROC(\%) for the transductive node classification on Wikipedia and Reddit.}
\label{tab: nc}
{
\begin{tabular}{c|cc|c}
\midrule
Methods    & Wikipedia    & Reddit       & Avg. Rank \\ \midrule
JODIE      & \textbf{88.99 ± 1.05} & 60.37 ± 2.58 & 5         \\
DyRep      & 86.39 ± 0.98 & 63.72 ± 1.32 & 5.5       \\
TGAT       & 84.09 ± 1.27 & \textbf{70.04 ± 1.09} & 4.5       \\
TGN        & 86.38 ± 2.34 & 63.27 ± 0.90 & 6.5       \\
CAWN       & 84.88 ± 1.33 & 66.34 ± 1.78 & 6         \\
TCL        & 77.83 ± 2.13 & \underline{68.87 ± 2.15} & 5.5       \\
GraphMixer & 86.80 ± 0.79 & 64.22 ± 3.32 & 5.5       \\
DyGFormer  & \underline{87.44 ± 1.08} & 68.00 ± 1.74 & \textbf{2.5}       \\
TGFormer   & 86.40 ± 2.24 & 66.41 ± 2.16 & \underline{4}  \\ \midrule
\end{tabular}
}
\end{table}
\subsection{Node Classification}
For the node classification task, we adhere to the evaluation protocols established by DyGFormer~\cite{yu2023towards}. The objective of this task is to predict the state label of the source node, given the node and future timestamps. Specifically, we employ the model obtained from the preceding transductive link prediction as the pre-trained model for node classification. A classifier decoder, such as a three-layer MLP, is then trained separately for the node classification task. We evaluate this task on two datasets with dynamic node labels, namely Wikipedia and Reddit, while excluding other datasets due to the absence of node labels.

Table~\ref{tab: nc} presents the results of our method compared to the baseline for the node classification task. Although our method does not achieve optimal performance, it remains competitive. The performance limitations can be attributed to the model’s emphasis on node interactions, such as periodic temporal pattern dependencies while placing less focus on individual node states, which adversely impacts its effectiveness in node classification.


\begin{figure*}[!ht]
\centering
    \subfigure{
        \includegraphics[width=0.31\linewidth]{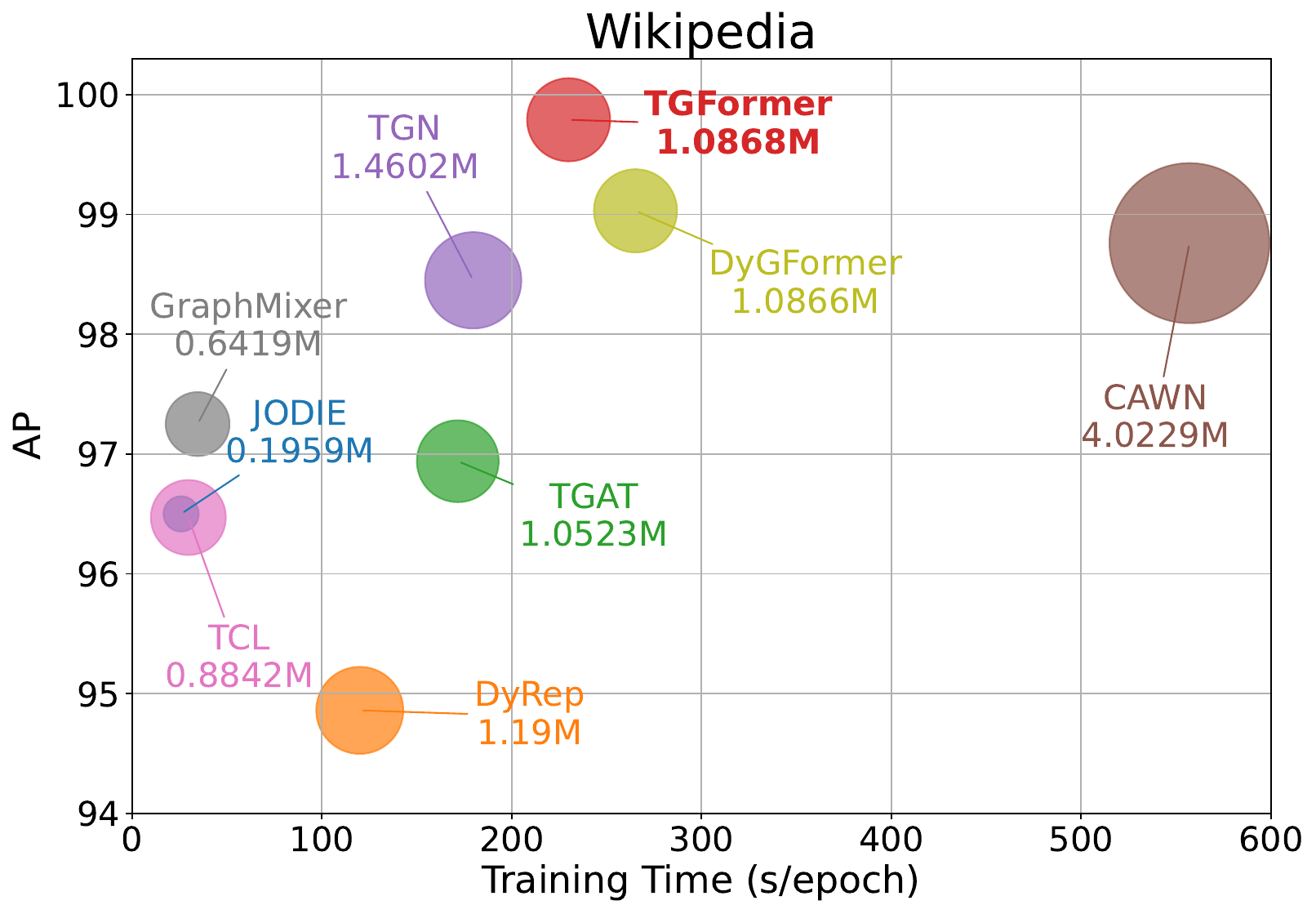}
    }\hspace{-3mm}
    \subfigure{
        \includegraphics[width=0.31\linewidth]{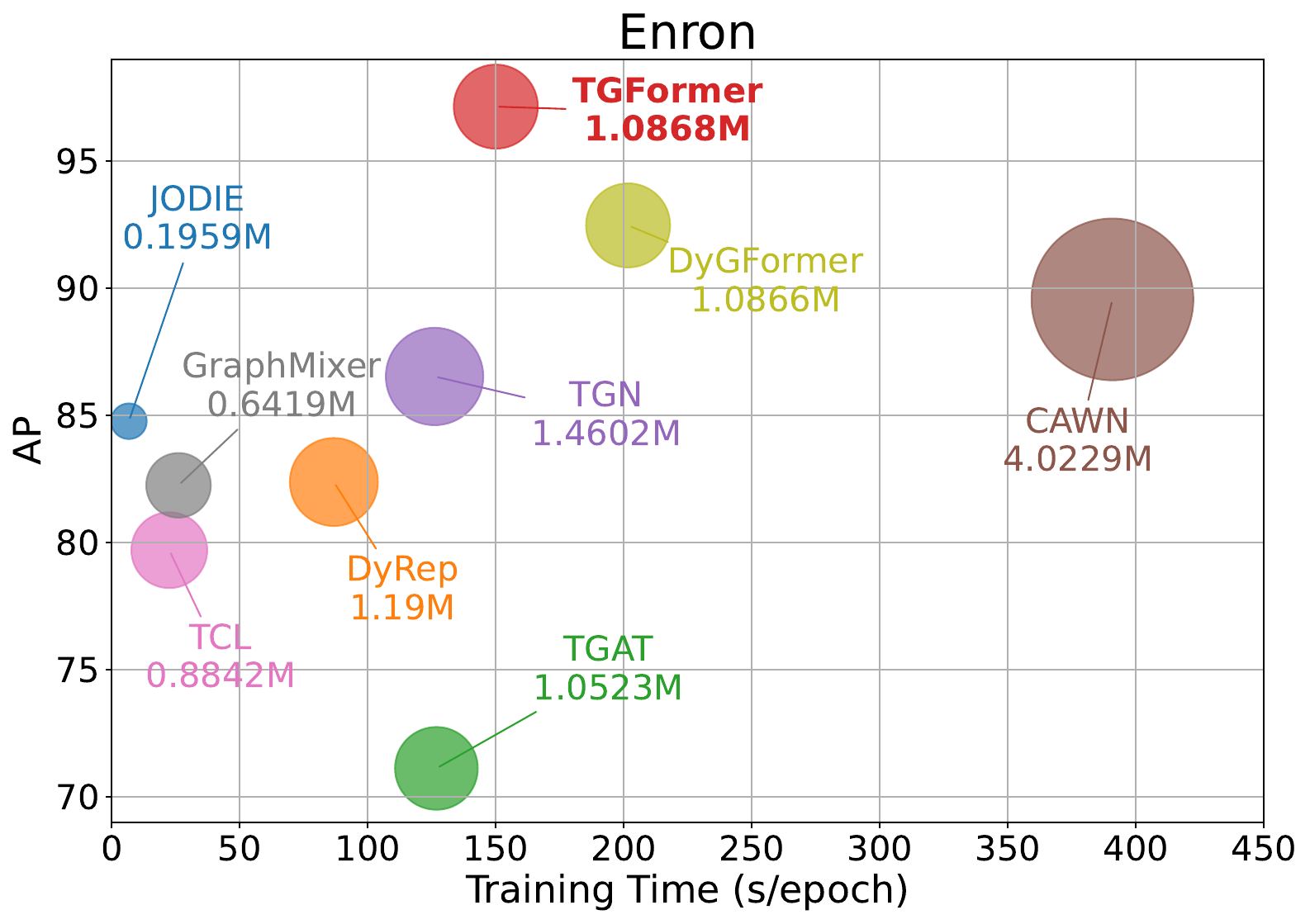}
    }\hspace{-3mm}
    \subfigure{
        \includegraphics[width=0.31\linewidth]{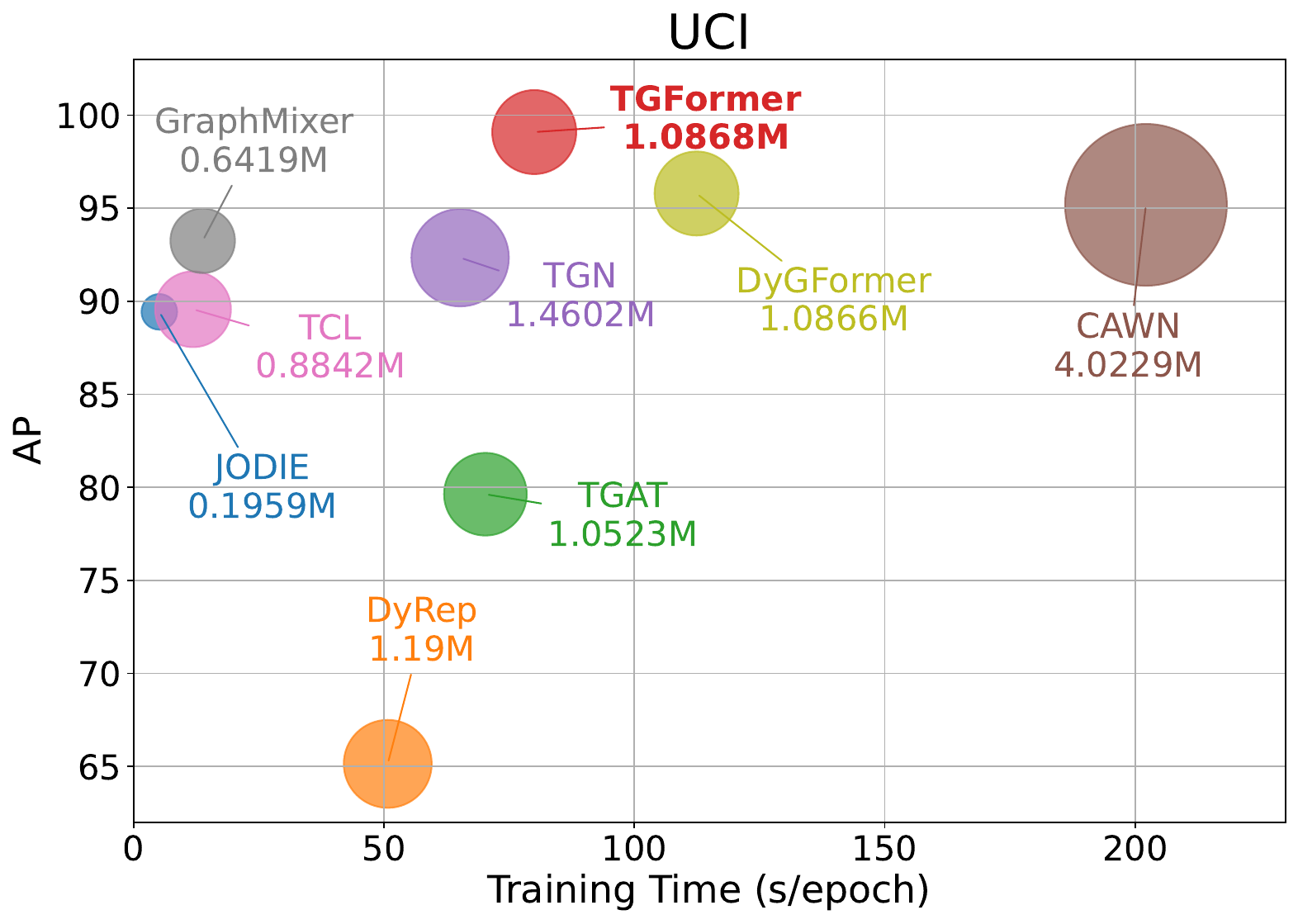}
    }
\caption{Comparison of model performance, parameter size and training time per epoch on Wikipedia, Reddit, and UCI.}
\label{fig: efficiency}
\end{figure*}

\subsection{Efficiency Analysis}
To compare the efficiency of the evaluated models, we conducted a comprehensive analysis encompassing performance in transductive link prediction using AP metrics, training time per epoch, and the number of trainable parameters between TGFormer and baseline methods across the Wikipedia, Enron, and UCI datasets. The results are depicted in Fig.~\ref{fig: efficiency}. It is evident that the walk-based method, such as CAWN, requires a longer training time due to its inefficient temporal walk operations and has a substantial number of parameters. Conversely, simpler and memory-based methods such as DyRep and JODIE may possess fewer parameters and exhibit faster training times; however, they exhibit a significant performance gap compared to the best-performing methods. In contrast, TGFormer outperforms the transformer-based method DyGFormer by achieving superior performance with a smaller size of trainable parameters and a moderate training time per epoch.

\begin{figure*}[!t]
\centering
    \subfigure[Wikipedia]{
        \includegraphics[width=0.32\linewidth]{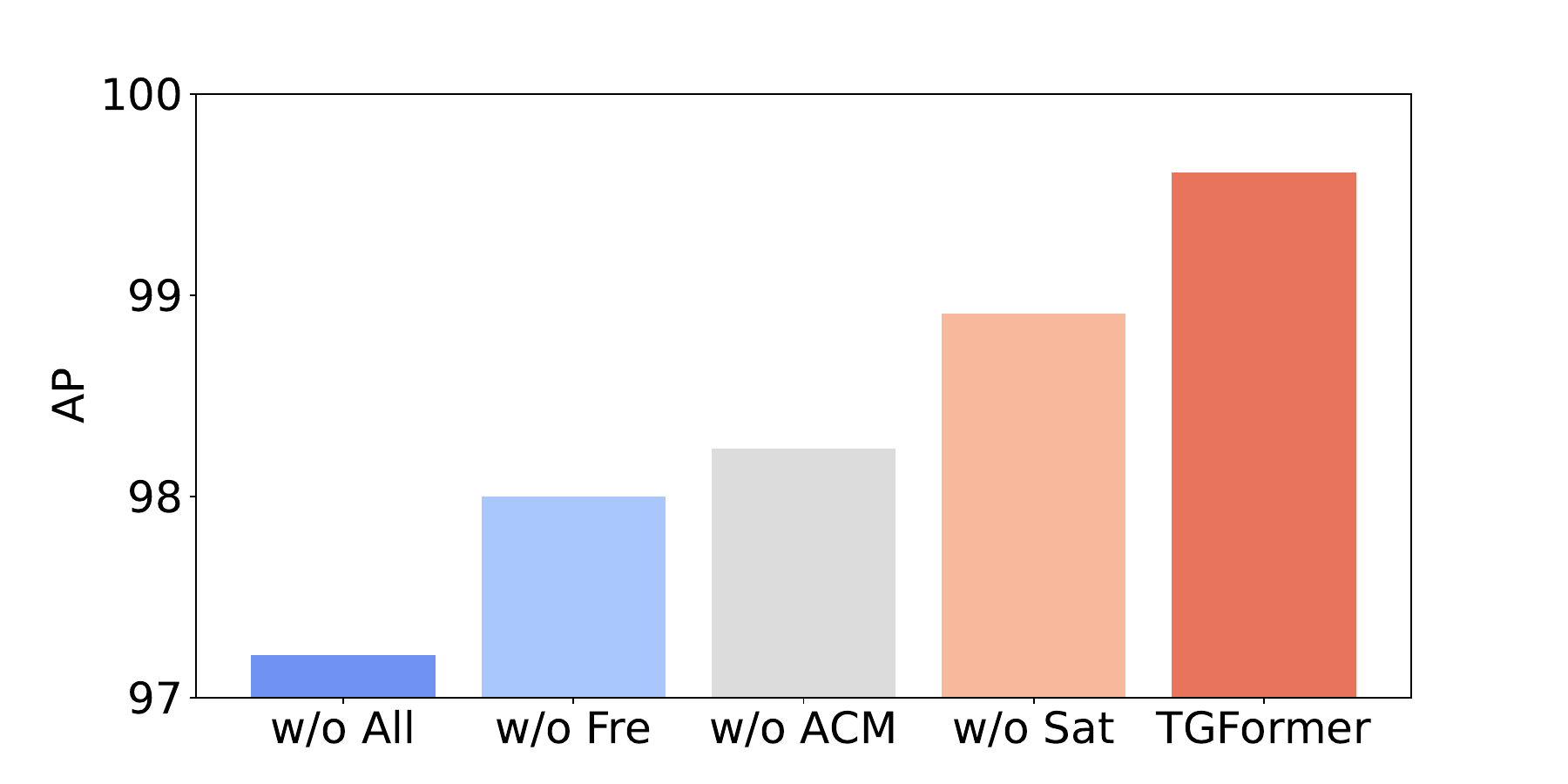}
    }\hspace{-6.5mm}
    \subfigure[Reddit]{
        \includegraphics[width=0.32\linewidth]{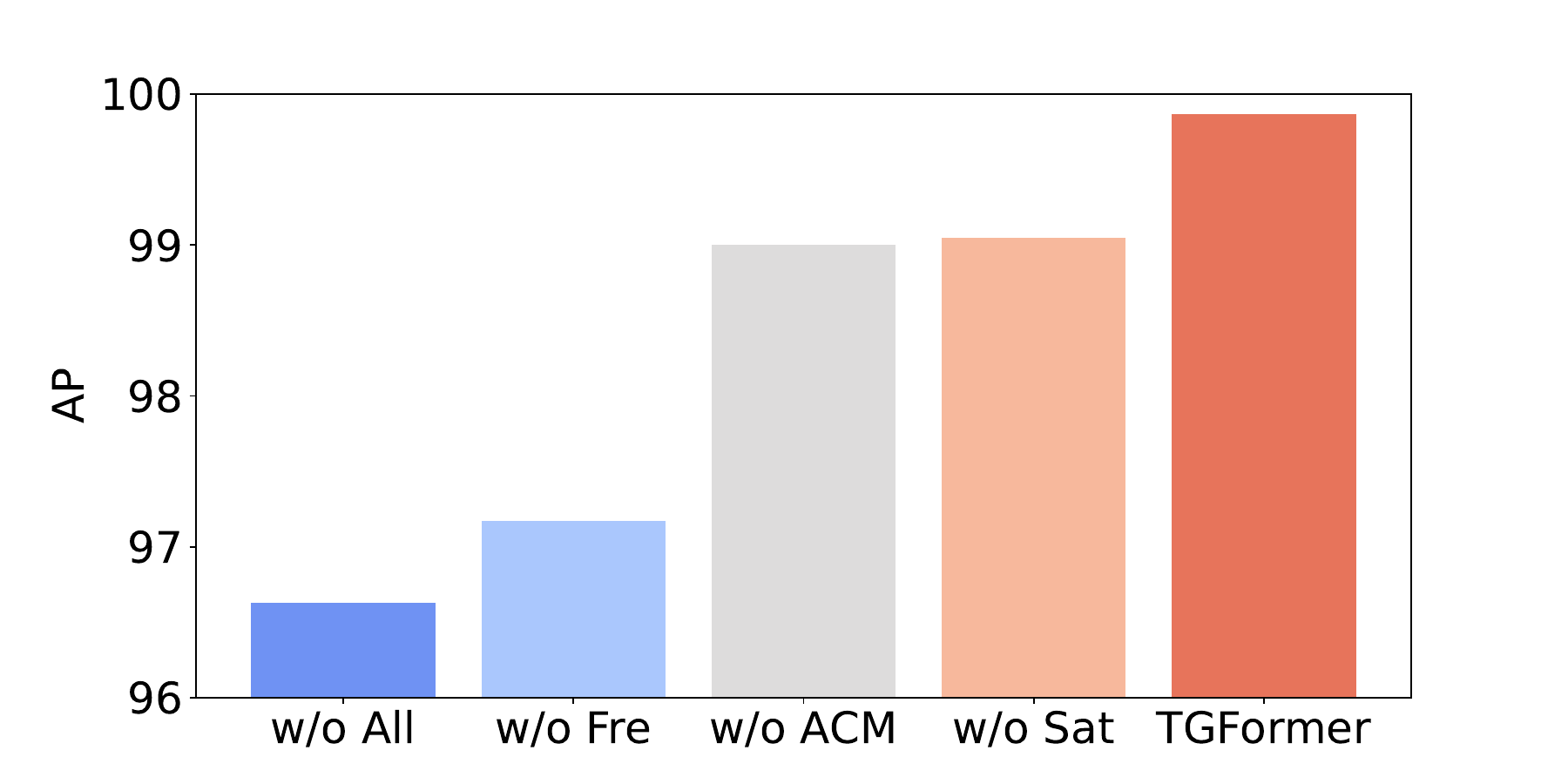}
    }\hspace{-6.5mm}
    \subfigure[LastFM]{
        \includegraphics[width=0.32\linewidth]{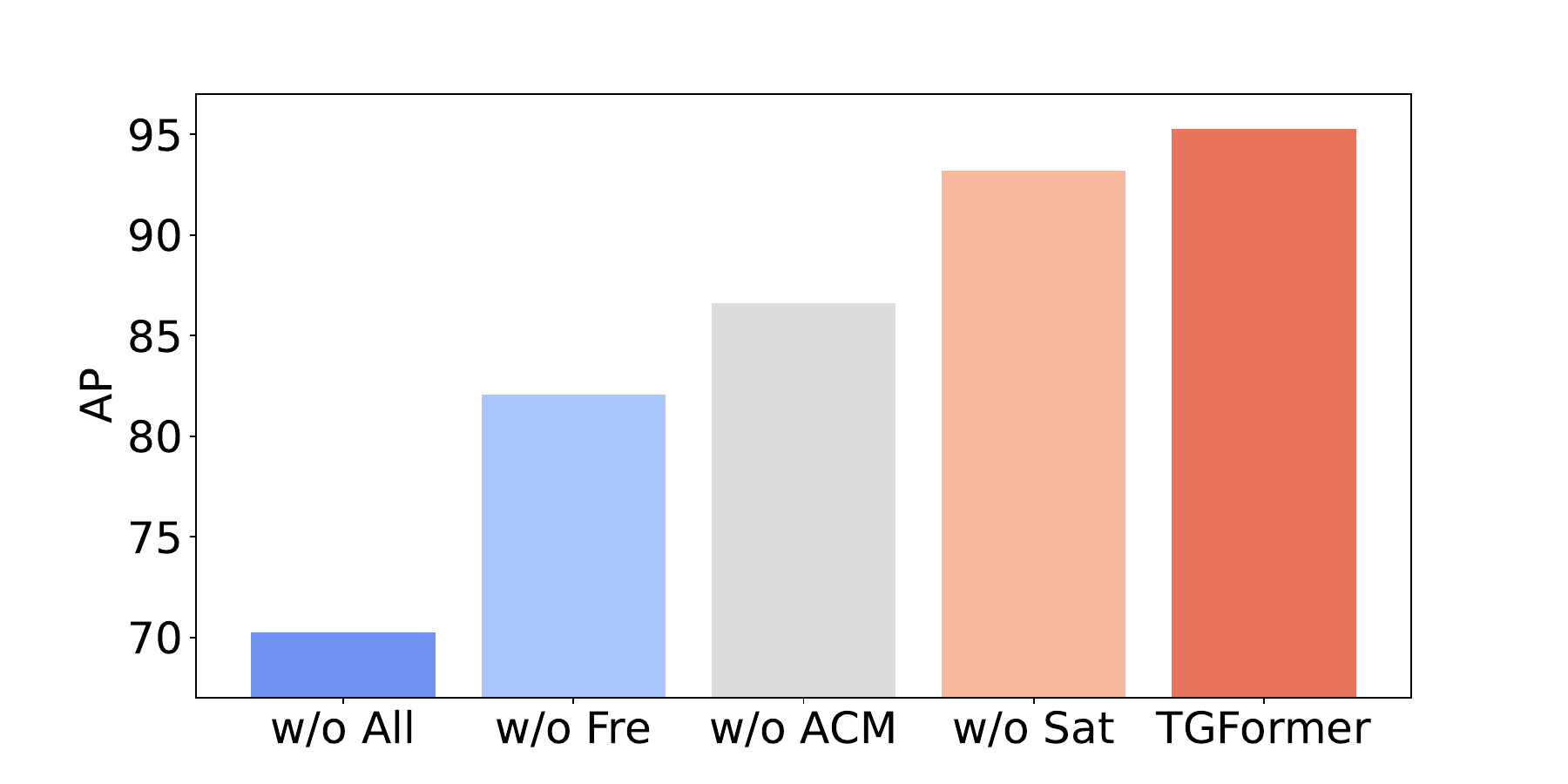}
    }
    \subfigure[Enron]{
        \includegraphics[width=0.32\linewidth]{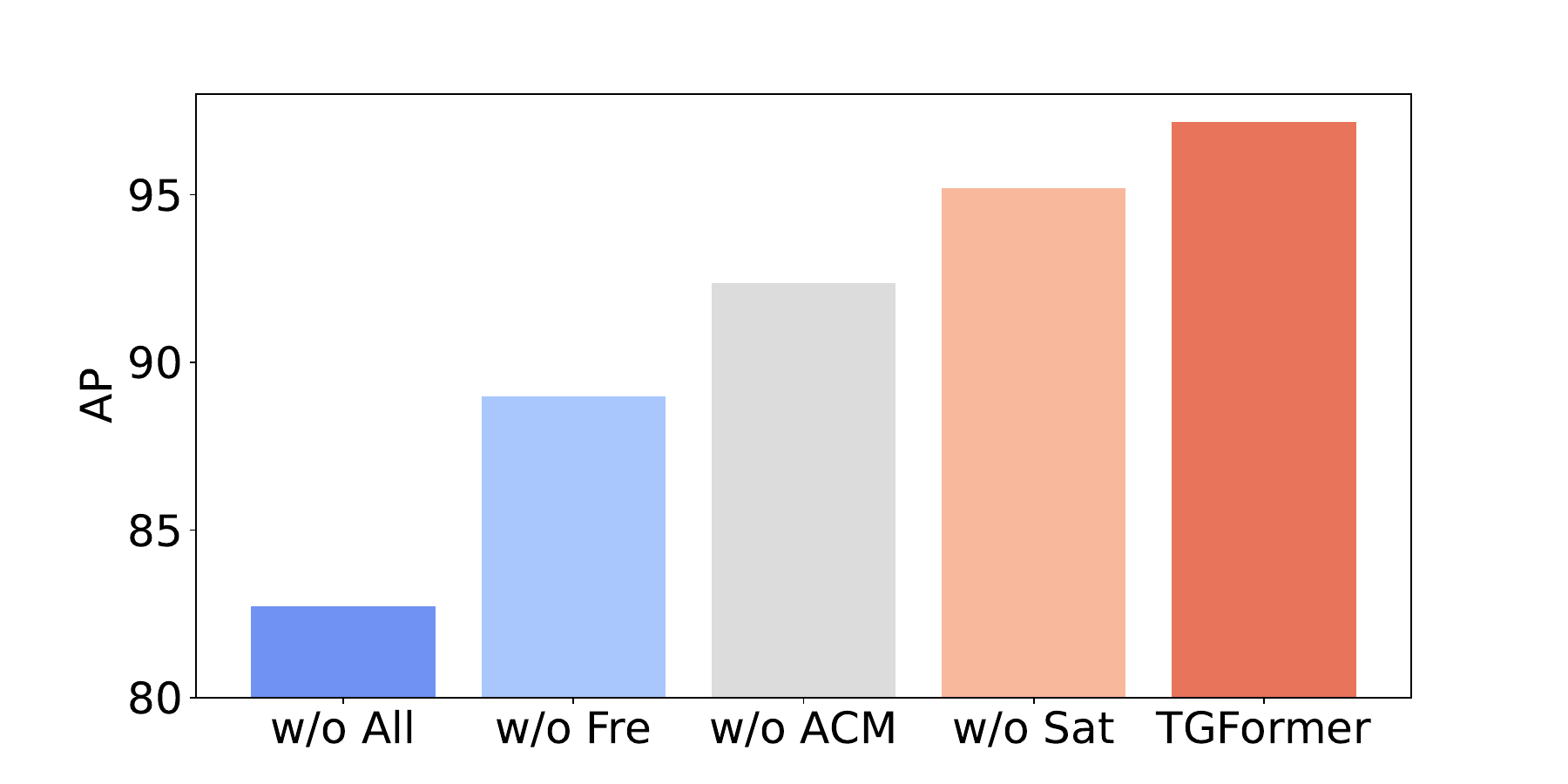}
    }\hspace{-6.5mm}
    \subfigure[UCI]{
        \includegraphics[width=0.32\linewidth]{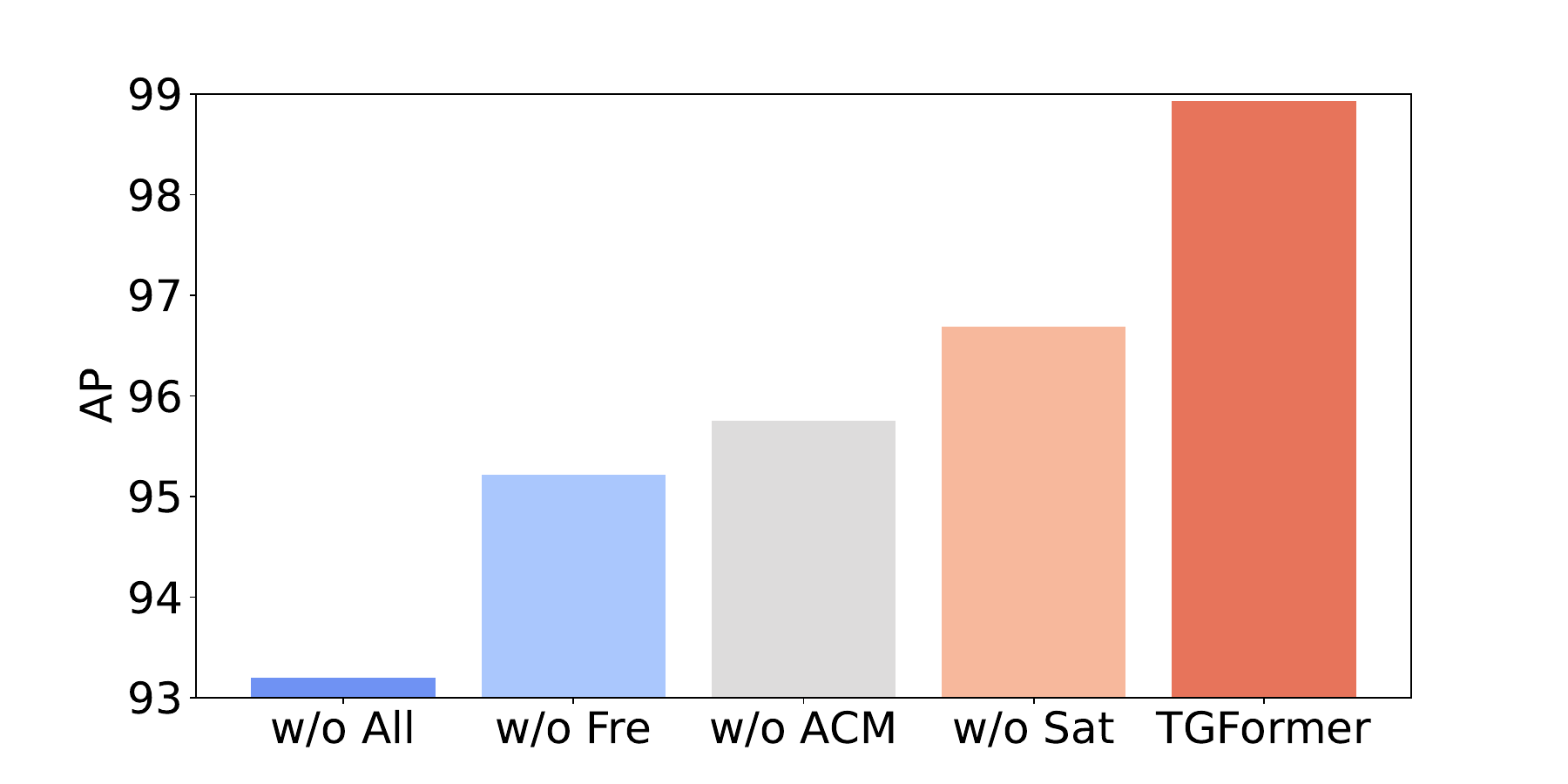}
    }\hspace{-6.5mm}
    \subfigure[CollegeMsg]{
        \includegraphics[width=0.32\linewidth]{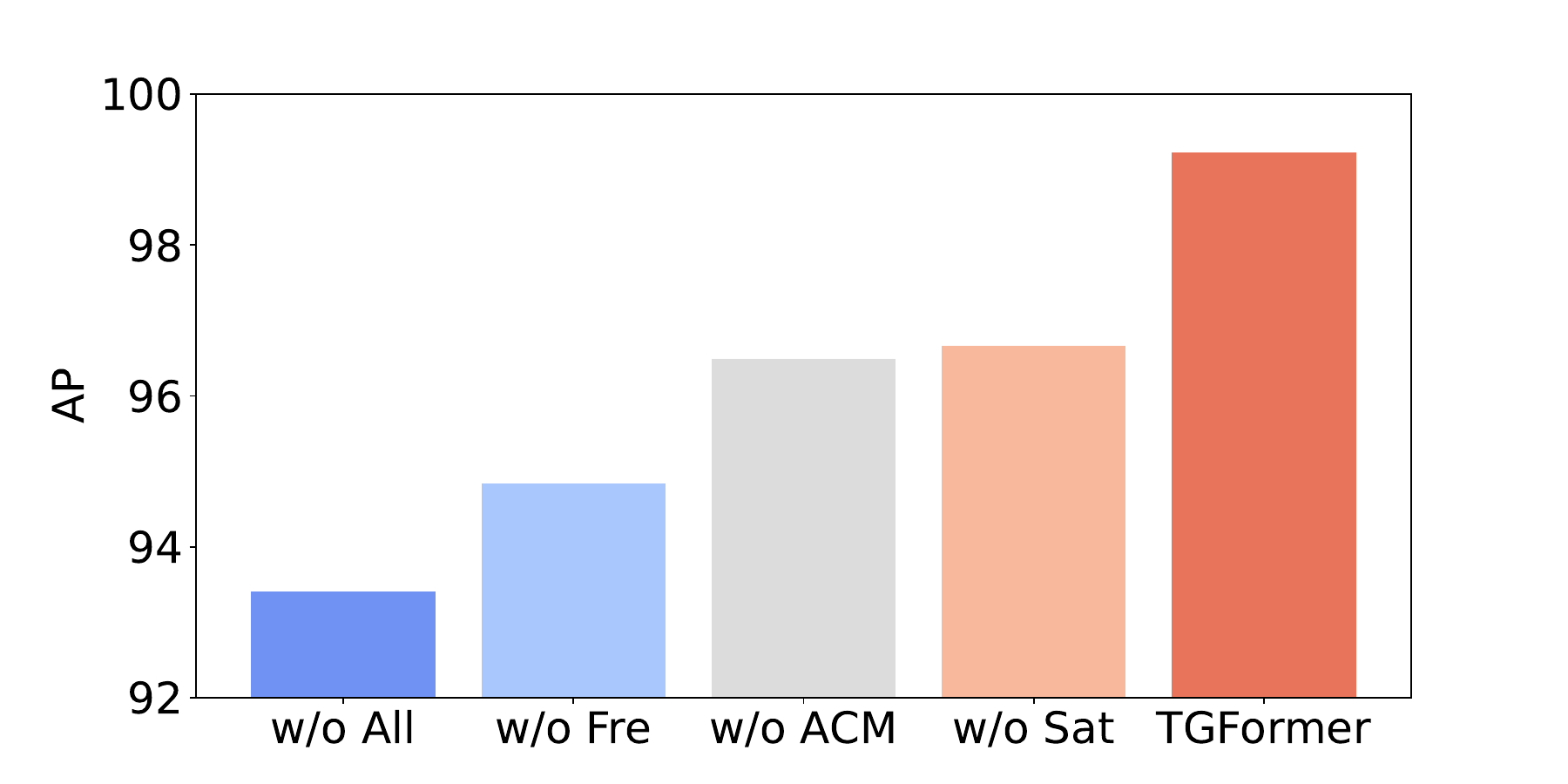}
    }
\caption{Ablation study in the transductive setting with the random negative sampling strategy. AP (\%) values are reported.}
\label{fig: ablation}
\end{figure*}
\subsection{Ablation Studies}
We further validate the effectiveness of modules in TGFormer through ablation studies, including frequency encoding (FE), auto-correlation mechanics (ACoM), and adaptive readout (AR). We refer to TGFormer without these three modules as w/o FE, w/o ACoM, and w/o AR respectively, and to TGFormer without all as w/o ALL. 
In detail, Specifically, in w/o FE, the encoding alignment module only takes node/link encoding and time encoding as input. In w/o ACoM, we replace the proposed ACoM with the traditional attention mechanics~\cite{vaswani2017attention}. In w/o AR, we replace the proposed adaptive readout with the traditional mean function~\cite{cong2023we}. We report the performance of the different modules on the six datasets in Fig.~\ref{fig: ablation}. We find that TGFormer typically performs best when all components are used, and the results are worse when any component is removed or changed. We can get the following conclusions: 1) removing the node interaction frequency encoding scheme significantly affects performance, as it directly captures the relationships between nodes and their interactions. 2) Although ACoM plays different roles on different datasets, on large datasets or datasets with significant periodicity, ACoM even plays a similar efficiency as FE. This suggests that ACoM can benefit from longer historical information and capture the periodic dependencies of the dataset. 3) The adaptive readout function can still achieve additional improvement.

\begin{table*}[!ht]
  \caption{TGFormer performance under different choices of hyper-parameter $c$ in the auto-correlation mechanism.}
  \label{tab: sensitivty}
  \centering
  \renewcommand{\multirowsetup}{\centering}
  \setlength{\tabcolsep}{5.0pt}
  \resizebox{\linewidth}{!}{
  \begin{tabular}{c|cc|cc|cc|cc|cc|cc}
    \toprule 
    Dataset & \multicolumn{2}{c}{Wikipedia}  & \multicolumn{2}{c}{Reddit}   & \multicolumn{2}{c}{LastFM} & \multicolumn{2}{c}{Enron} & \multicolumn{2}{c}{UCI} & \multicolumn{2}{c}{CollegeMsg}  \\
    \cmidrule(lr){2-3} \cmidrule(lr){4-5}\cmidrule(lr){6-7} \cmidrule(lr){8-9}\cmidrule(lr){10-11}\cmidrule(lr){12-13}
    Metric & AP & AUC & AP & AUC & AP & AUC & AP & AUC & AP & AUC & AP & AUC  \\
    \midrule
$c=1$ & 99.61 & 99.58 & 99.87 & 99.83 &  96.26 & 96.55& 97.17 & 97.24 & 98.93 & 98.35 & 99.23 & 98.95 \\
$c=2$ & \textbf{99.97} & \textbf{99.97} & 99.88 & 99.84  & 96.28 & 96.54& 96.62 & 96.68 & 99.02 & 98.65 & \textbf{99.29} & \textbf{99.09} \\
$c=3$ & 99.86 & 99.83 & \textbf{99.89} & \textbf{99.86}  & 96.34 & 96.64& \textbf{97.85} & \textbf{97.94} & 99.22 & 98.79 & 99.18 & 98.75 \\
$c=4$ & 99.58 & 99.57 & 99.87 & 99.82  & 96.44 & 95.70 & 97.56 & 97.72 & 99.24 & 98.82 & 99.28 & 98.94 \\
$c=5$ & 99.93 & 99.93 & 99.88 & 99.85  & \textbf{96.63} & \textbf{96.50} & 97.73 & 97.84 & \textbf{99.27} & \textbf{98.92} & 99.30  & 98.97    \\ 
    \bottomrule
  \end{tabular}
  }
\end{table*}

\subsection{Parameter Sensitivity Analysis}
As demonstrated in Table~\ref{tab: sensitivty}, we can confirm the model's robustness concerning the hyper-parameter $c$ (Eq.~\eqref{eq: topk}). To achieve an optimal balance between performance and computational efficiency, we set $c$ to the range of 1 to 5. It was also observed that the effects of most of the data sets did not fluctuate significantly and were more stable, all being somewhat periodic.

\begin{figure}[!ht]
\centering
\includegraphics[width=0.85\linewidth]{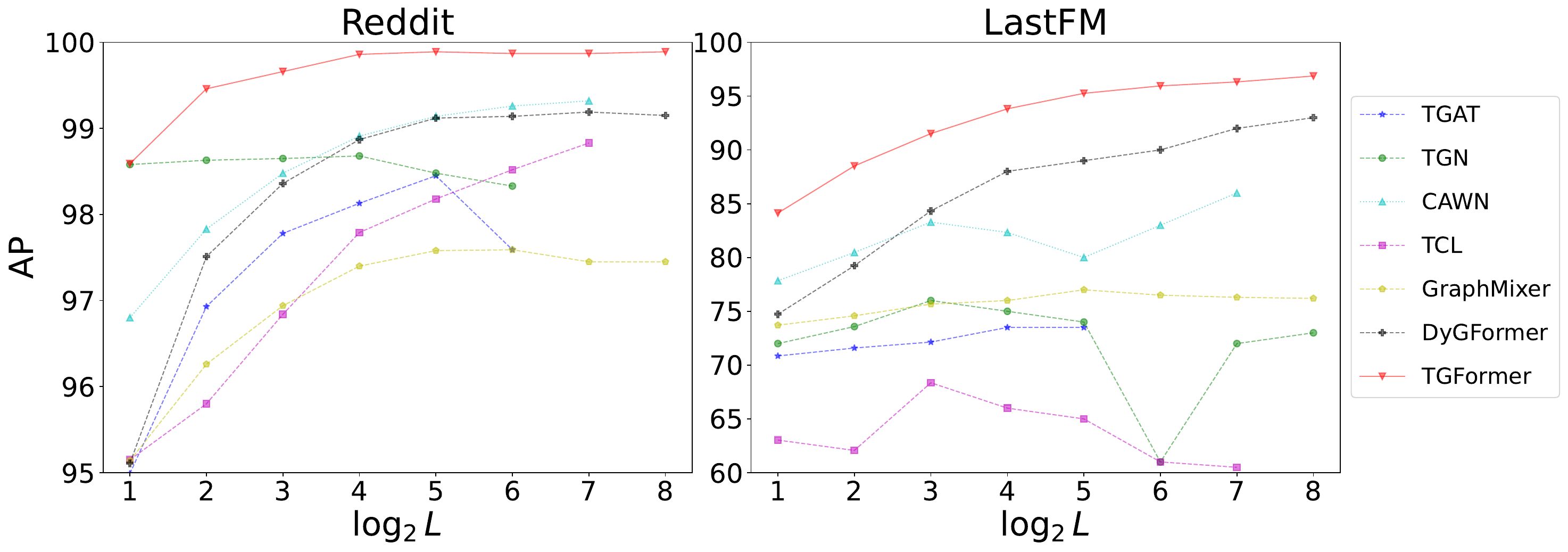}
\caption{The performance of various methods on Reddit and LastFM datasets across different historical lengths $L$.}
\label{fig: Long_term}
\end{figure}

\begin{figure}[!ht]
\centering
\includegraphics[width=0.85\linewidth]{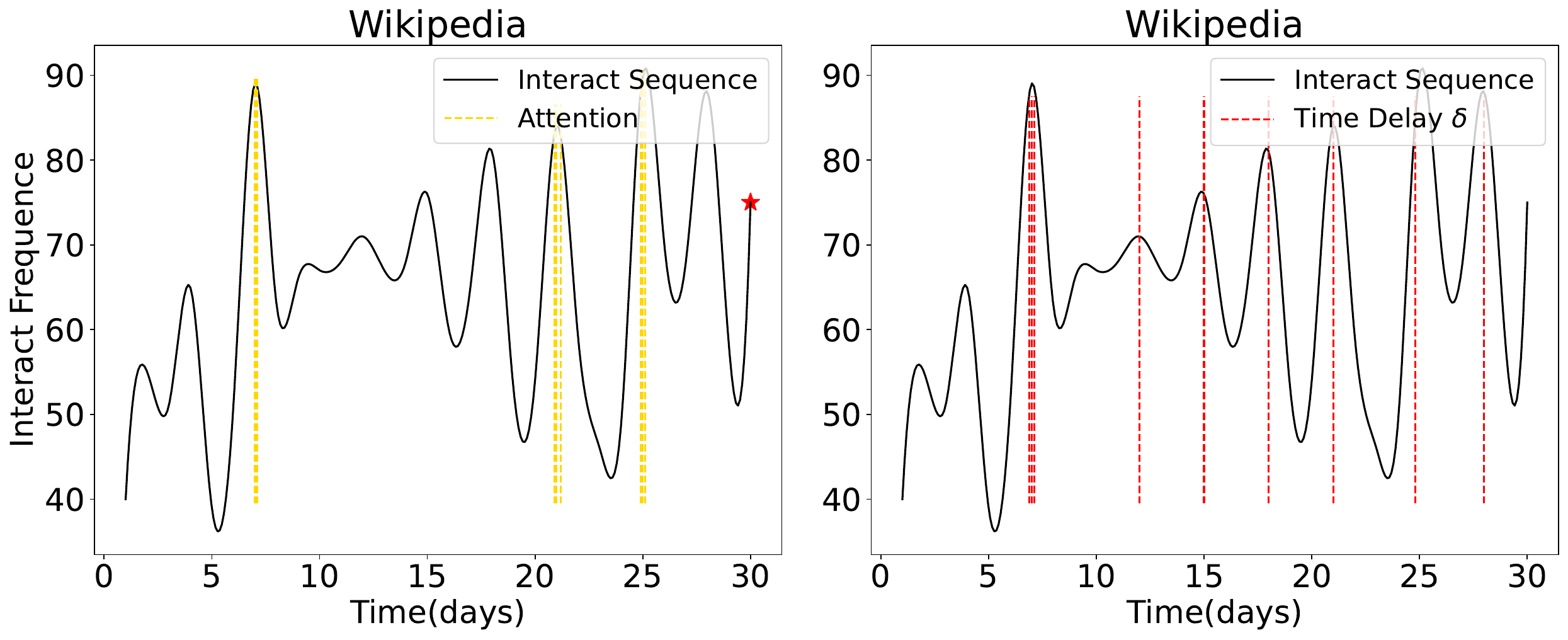}
\caption{
Visualization of learned periodic dependencies. For clearness, we select the top-9 time delay sizes $\left\{\delta_1,\cdots,\delta_9, \right\}$ of Auto-Correlation and mark them in raw series (red lines). For attention, top-9 similar points with respect to the last time step (red stars) are also marked by yellow lines.}
\label{fig: Periodic}
\end{figure}

\begin{figure}[!ht]
\centering
    \subfigure[Orignal Graph]{
        \includegraphics[width=0.32\linewidth]{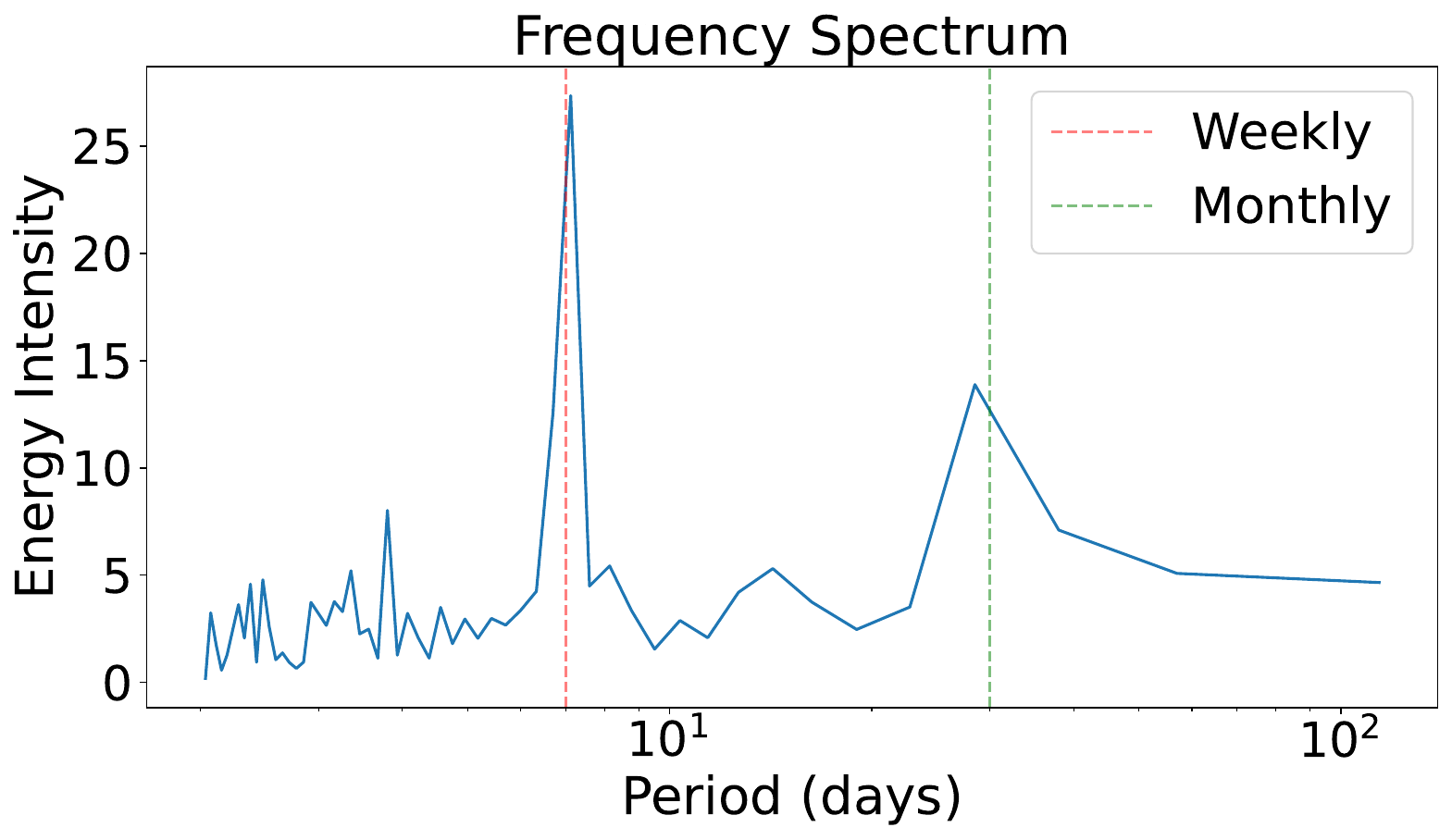}
    }\hspace{-1.5mm}
    \subfigure[DyGFormer]{
        \includegraphics[width=0.32\linewidth]{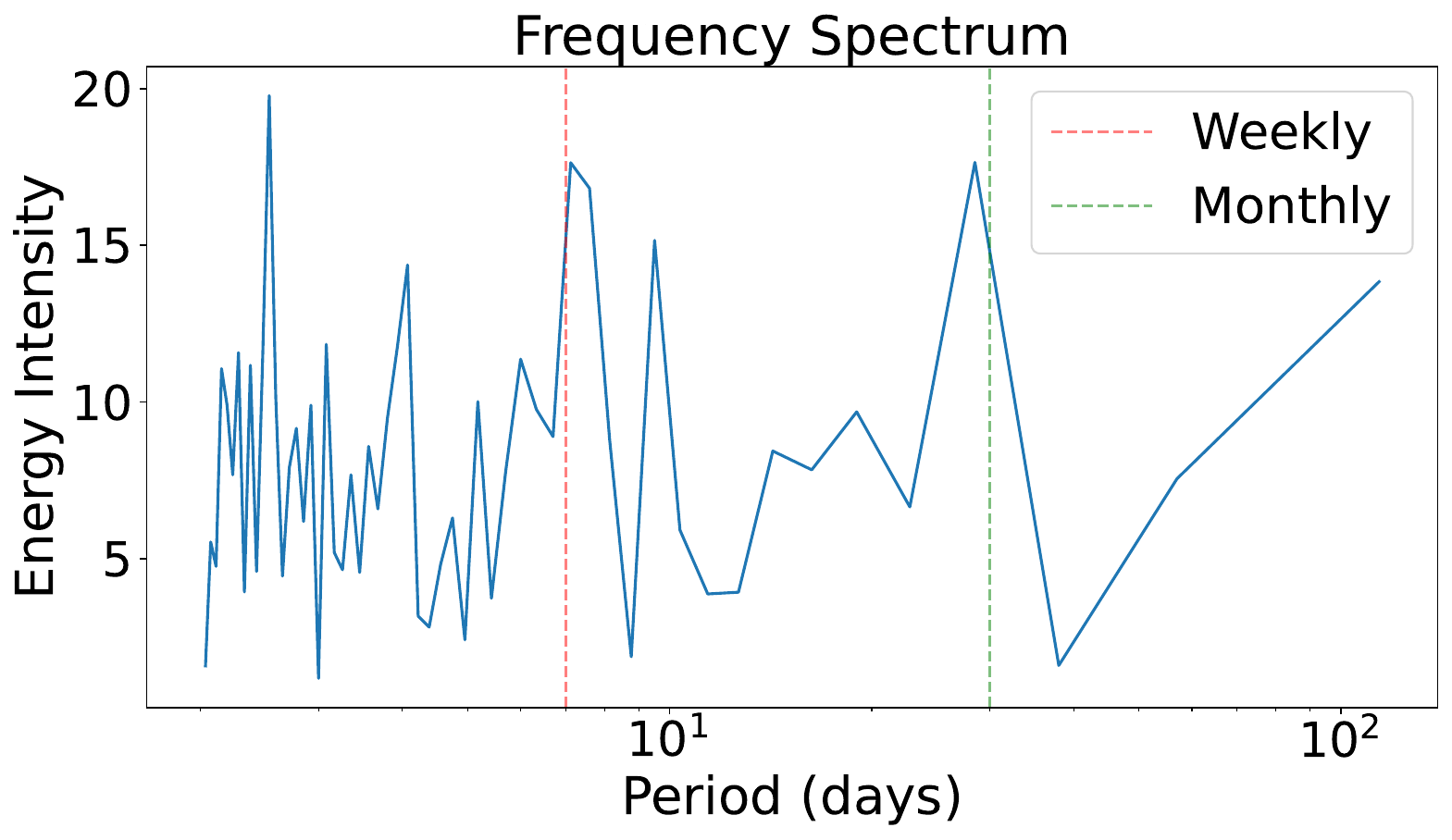}
    }\hspace{-1.5mm}
    \subfigure[TGFormer]{
        \includegraphics[width=0.32\linewidth]{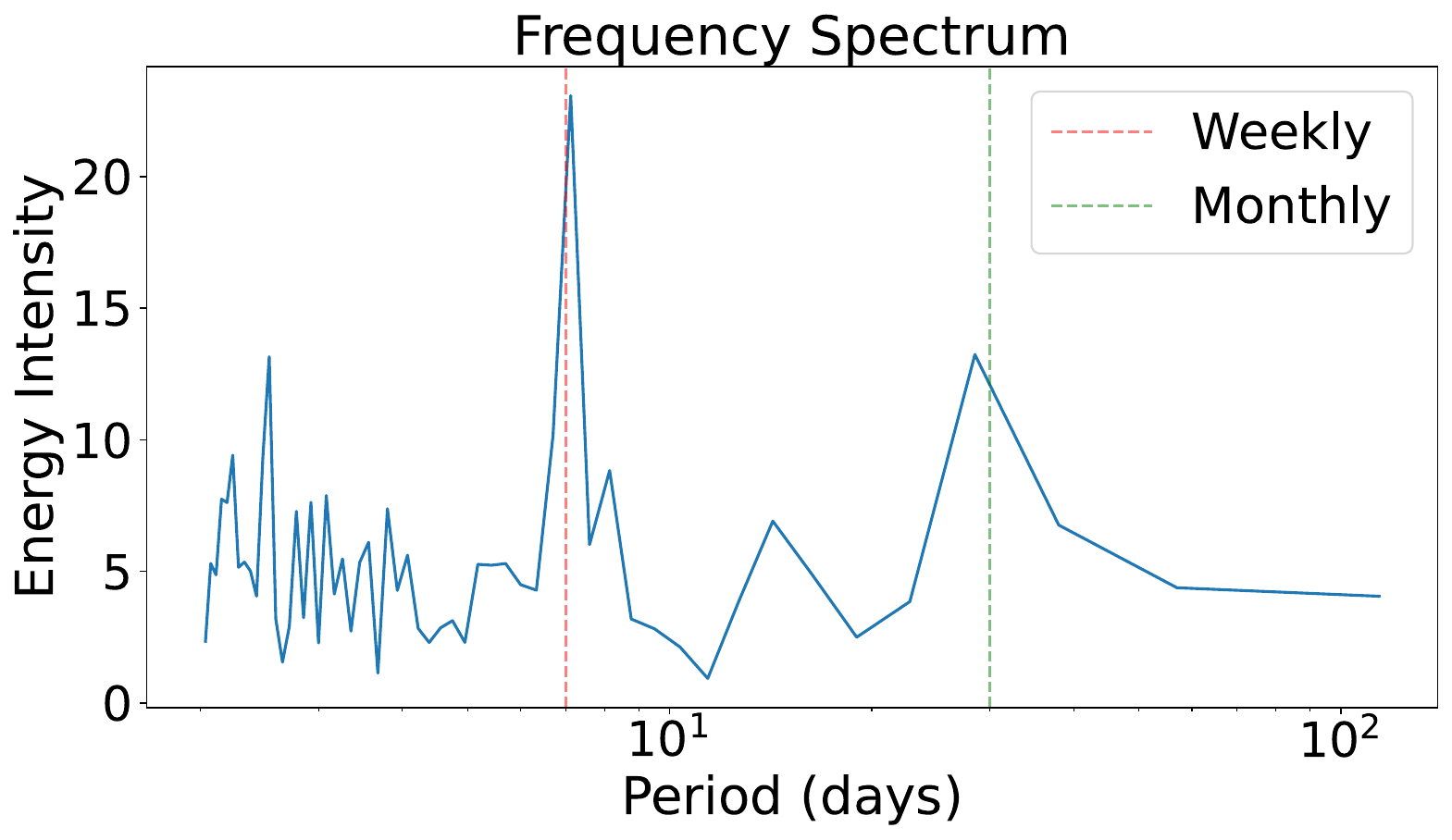}
    }
\caption{The performance of different methods Original Graph(left), DyGFormer (middle), and TGFormer (right) in capturing periodic temporal patterns.}
\label{fig: Case_study}
\end{figure}


\subsection{Long-term Temporal Dependency}
We validate the advantages of our transformer architecture in effectively and efficiently utilizing longer histories. Our experiments are conducted on the Reddit and LastFM datasets, chosen for their potential to benefit from long-term historical data. For baseline comparisons, we augment the sampling of neighbors or increase the number of causal anonymous walks (starting from 2) to empower them with access to extended histories. The experimental results are illustrated in Fig.~\ref{fig: Long_term}, with the x-axis depicted on a logarithmic scale with a base of 2. It is noteworthy that some baseline results are incomplete due to out-of-memory errors encountered at longer historical lengths. For instance, CAWN reaches an out-of-memory state when the historical length extends to 256. From Fig.~\ref{fig: Long_term}, we deduce the following: 1) a majority of the baselines exhibit deteriorated performance with increasing historical lengths, indicative of their limitations in capturing long-term temporal dependencies; 2) the baselines generally incur substantial computational costs when processing extended histories. While memory network-based methods (e.g., TGN) manage to handle longer histories with reasonable computational costs, they do not significantly benefit from these extended histories due to issues such as vanishing or exploding gradients; 3) TGFormer consistently demonstrates performance gains with longer historical sequences, underscoring the superiority of the transformer architecture in effectively utilizing extended historical records.

\subsection{Auto-Correlation vs. attention}
We also verified the advantage of the ACoM in capturing periodic temporal dependencies. We conducted experiments on Wikipedia to determine the model's effectiveness in capturing periodicity by measuring the interaction frequency of one of the nodes. For clarity, we visualized the learned periodic dependency in Fig~\ref{fig: Periodic}, where the x-axis represents the node interaction time divided by days, and the y-axis denotes the node interaction frequency during this period. For comparison with attention mechanisms, the top -9 time delay sizes $\left\{\delta_1,\cdots,\delta_9, \right\}$ of ACoM are highlighted within the raw series using red lines. In contrast, the top-9 most similar data points relative to the final time step, as determined by attention mechanisms, are denoted by yellow lines with red stars. As illustrated, while attention mechanisms capture node interaction intent, they fail to capture interaction periodicity effectively. This demonstrates that our model can discover relevant information more comprehensively and accurately.

To more intuitively demonstrate TGFormer's ability to capture periodicity, we utilize a dataset generated by the Dynamic Stochastic Block Model (DSBM). This dataset consists of 1,000 nodes, 24,430 edges, and spans one year with timestamps as the temporal granularity, exhibiting both weekly and monthly periodicity. Specifically, we provide the first 70\% of the data to DyGFormer (attention mechanism) and TGFormer (auto-correlation mechanism) to generate the remaining 30\%. By applying Fourier transform analysis to detect periodicity in the generated data, we validate TGFormer's capability to capture the inherent periodicity of time-evolving data. Our results, presented in Fig.~\ref{fig: Case_study}, illustrate this capability: (a) represents the original generated graph data, which exhibits distinct weekly and monthly periodic patterns; (b) shows the graph data generated by DyGFormer; and (c) presents the graph data generated by TGFormer. These findings suggest that transforming data from the time domain to the frequency domain allows ACoM to more effectively identify key periodic patterns and correlations that may be difficult to detect in the time domain. In contrast, directly analyzing the time domain may fail to fully capture signal periodicity, particularly when the data contains noise or multiple overlapping frequency components. The frequency domain representation enables a clearer separation of different signal frequencies, thereby enhancing the identification of relevant patterns, especially for signals exhibiting periodic behavior.

\begin{figure}[!ht]
\centering
\includegraphics[width=0.5\linewidth]{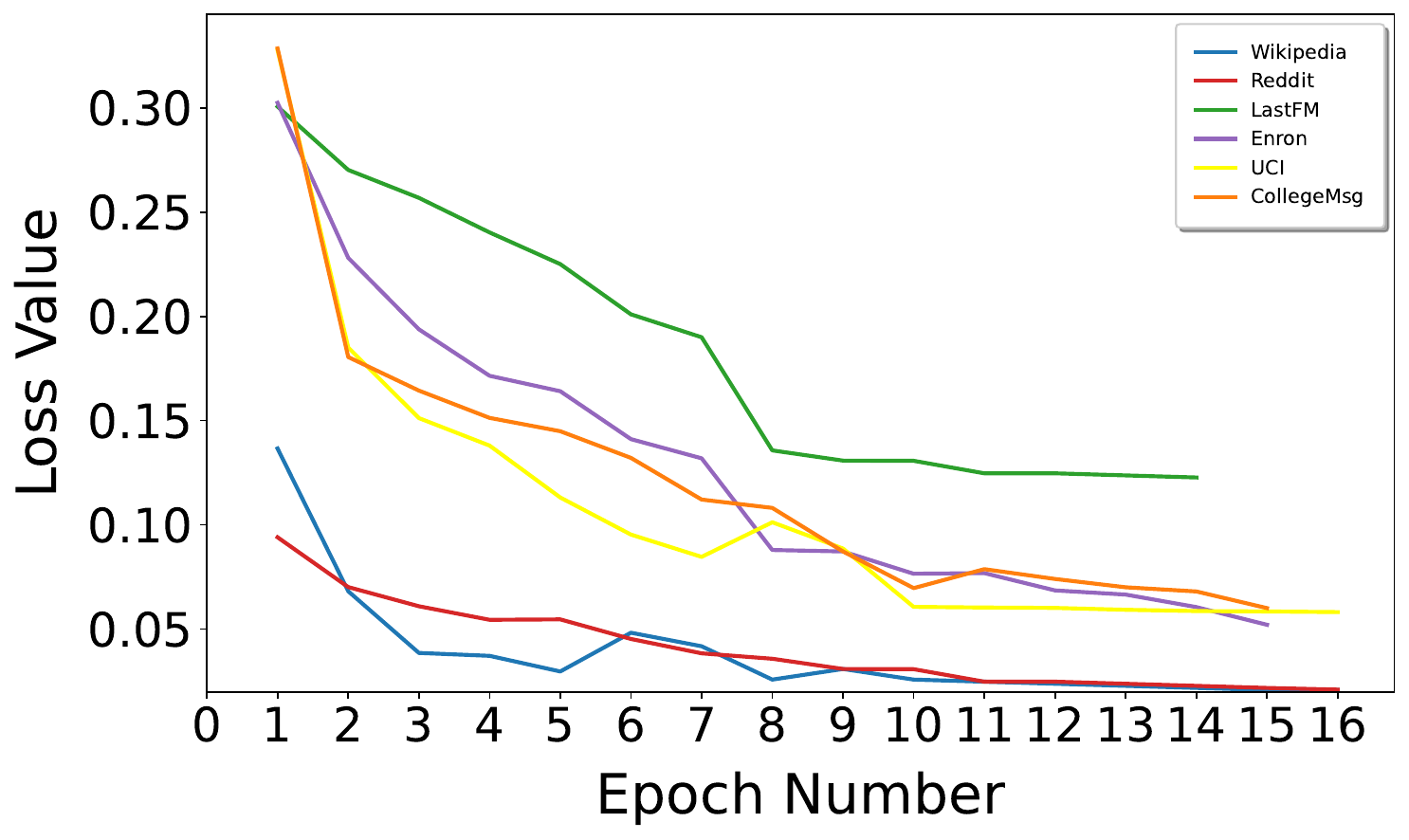}
\caption{The performance of TGFormer loss value on various datasets across epoch numbers.}
\label{fig: Loss}
\end{figure}

\begin{figure*}[!ht]
\centering
\includegraphics[width=0.95\linewidth]{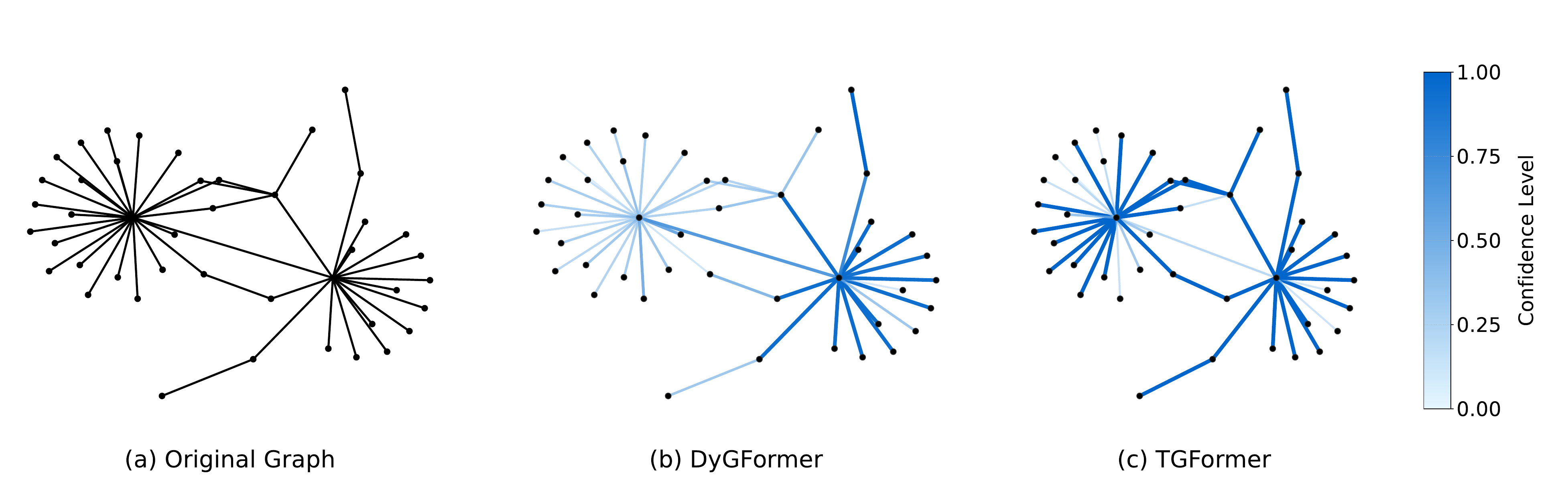}
\caption{Visualization of DyGFormer and TGFormer on link prediction performance.}
\label{fig: link}
\end{figure*}

\subsection{Case study for link prediction}
To illustrate the superiority of TGFormer more intuitively, we visualize the loss optimization process, as shown in Fig.~\ref{fig: Loss}. TGFormer's loss converges within 20 epochs across all datasets. By analyzing the convergence trends, we observe that datasets with a larger number of nodes exhibit faster convergence and lower corresponding loss values. This suggests that a greater number of node samples in each training iteration enhances the model's learning capability. Furthermore, we visualize TGFormer's advantages in link prediction, as shown in Fig.~\ref{fig: link}, where (a) represents the original graph, (b) shows the prediction results of DyGFormer, and (c) presents the prediction results of TGFormer. The darkness of each edge indicates the probability assigned by the model to the existence of that edge. The results demonstrate that TGFormer outperforms DyGFormer by achieving more accurate link predictions.

\section{Conclusion}
In this paper, we introduce TGFormer, a novel approach that redefines temporal graph learning by framing it within the context of time series analysis. Our model pioneers an auto-correlation mechanism that surpasses conventional attention-based methods, enabling the effective capture of long-term temporal dependencies and periodic patterns. This innovation not only enhances computational efficiency but also optimizes information utilization, making TGFormer well-suited for complex temporal graph tasks. Comprehensive experiments on diverse real-world datasets validate the model's effectiveness and efficiency, demonstrating its superior performance in temporal representation learning.

Despite its promising results, TGFormer has certain limitations. Notably, its reliance on one-hop neighbors restricts its ability to capture high-order relationships, which may hinder performance in scenarios requiring deeper relational modeling. Furthermore, its computational scalability remains a challenge, particularly for extremely large datasets. To address these limitations, future research will explore the integration of multi-hop neighbor aggregation to enrich the relational context and enhance model expressiveness. Additionally, we aim to investigate dynamic and scalable architectures capable of supporting large-scale data parallelism, a critical requirement for industrial applications that demand high efficiency and adaptability.

\section*{Acknowledgment}
This work was supported in part by the Zhejiang Provincial Natural Science Foundation of China under Grant LDT23F01012F01, in part by the National Natural Science Foundation of China under Grants 62372146 and 62202304.

\bibliographystyle{elsarticle-num} 
\bibliography{ref}

\clearpage
\appendix
\section{Detail descriptions of datasets}
\label{sec: dataset}
Here, we briefly introduce the mechanisms of these methods for our assessment. We consider the following baselines:
\begin{itemize}
    \item \textbf{Wikipedia}\footnote{http://snap.stanford.edu/jodie/wikipedia.csv} comprises a bipartite interaction graph that chronicles user edits on Wikipedia pages over a one-month period. In this dataset, nodes represent both users and pages, while links denote editing actions, each associated with a timestamp and a 172-dimensional LIWC (Linguistic Inquiry and Word Count) feature. Additionally, dynamic labels are provided to indicate users who faced temporary bans from the platform.
    \item \textbf{Reddit}\footnote{http://snap.stanford.edu/jodie/reddit.csv} consists of a bipartite graph monitoring user posts on Reddit for one month. Here, nodes represent users and subreddits, whereas links denote timestamped posting requests, each accompanied by a 172-dimensional LIWC feature. This dataset also includes dynamic labels indicating whether users were banned from posting.
    \item \textbf{LastFM}\footnote{http://snap.stanford.edu/jodie/lastfm.csv} contains a bipartite dataset that records user interactions with songs over a one-month period. Nodes represent users and songs, while links denote the listening behaviors of users.
    \item \textbf{Enron}\footnote{https://www.cs.cmu.edu/~./enron/} documents email communications among employees of the ENRON energy corporation over a three-year period. This dataset does not include node attributes or link features.
    \item \textbf{UCI}\footnote{http://konect.cc/networks/opsahl-ucforum/} comprises an online communication network where nodes represent university students, and links denote messages exchanged among them. This dataset is particularly pertinent for scholarly examinations of virtual interactions within student communities.
    \item \textbf{CollegeMsg}\footnote{https://snap.stanford.edu/data/CollegeMsg.html} represents an online social network at the University of California, depicting user interactions through private messages exchanged at various timestamps. This dataset does not include node labels or edge features.
\end{itemize}

\section{Detail descriptions of baselines}
\label{sec: baseline}
Here, we briefly introduce the mechanisms of these methods for our assessment. We consider the following baselines:
\begin{itemize}
    \item \textbf{JODIE}~\cite{kumar2019predicting} utilizes two interconnected recurrent neural networks to update user and item states, enhancing precision in capturing intricate temporal patterns within user-item interactions. The incorporation of a projection operation enables accurate learning of future representation trajectories. 
    \item \textbf{DyRep}~\cite{trivedi2019dyrep} introduces a recurrent architecture that systematically updates node states after each interaction within temporal graphs. Furthermore, it incorporates a temporal-attentive aggregation module, enabling the model to consider the evolving structural information in temporal graphs over time. This dual-component framework provides a consistent and effective approach to capture intricate temporal dynamics and evolving graph structures in dynamic networks.
    \item \textbf{TGAT}~\cite{xu2020inductive} utilizes self-attention to compute node representations, aggregating features from temporal neighbors and employing a time encoding function for capturing nuanced temporal patterns. This concise approach ensures precise analysis of evolving graph structures and temporal dynamics in dynamic networks.
    \item \textbf{TGN}~\cite{rossi2020temporal} utilizes an evolving memory system for each node, updated upon node interactions through the message function, message aggregator, and memory updater mechanisms. Simultaneously, an embedding module is employed to generate temporal node representations, ensuring a dynamic and comprehensive analysis of evolving graph structures in complex networks.
    \item \textbf{CAWN}~\cite{wang2021inductive}  initiates its process by extracting multiple causal anonymous walks for each node, facilitating an in-depth examination of the network dynamics' causality and the establishment of relative node identities. Following this, RNN is utilized to encode each individual walk. The encoded walks are subsequently aggregated to synthesize the final node representation. 
    \item \textbf{EdgeBank}~\cite{poursafaei2022towards} stands as a purely memory-based approach for transductive temporal link prediction, distinguished by its absence of trainable parameters. The method operates by storing observed interactions within a dedicated memory unit, which is then meticulously updated using diverse strategies. This consistent reliance on memory and strategic updates forms the core of EdgeBank, ensuring a robust and parameter-free framework for precise and transductive predictions in temporal link prediction tasks.
    \item \textbf{TCL}~\cite{wang2021tcl} initiates its methodology by generating interaction sequences for each node via a breadth-first search algorithm applied to the temporal dependency interaction sub-graph. Subsequently, it employs a graph transformer that simultaneously considers graph topology and temporal information to derive node representations. Furthermore, TCL integrates a cross-attention mechanism to model the complex interdependencies between paired interaction nodes effectively.
    \item \textbf{GraphMixer}~\cite{cong2023we} adopts a fixed time encoding function, surpassing its trainable counterpart in performance. It incorporates this function into an MLP-Mixer-based link encoder to learn from temporal links efficiently. The framework utilizes neighbor mean-pooling in the node encoder for a concise summarization of node features.
    \item \textbf{DyGFormer}~\cite{yu2023towards} introduces a Transformer-based architecture enhanced by a neighbor co-occurrence coding scheme, effectively capturing node correlations within interactions. Also, it employs patching techniques to enable the model to capture long-term temporal dependencies. 
    \item \textbf{FreeDyG}~\cite{tian2024freedyg}: FreeDyG incorporates a node interaction frequency encoding module that captures the proportion of common neighbors and the frequency of interactions between node pairs. Shifting the focus to the frequency domain, it allows the model to better capture periodic patterns and dynamic shifts in node interactions over time.
\end{itemize}


\section{Detailed descriptions of experimental setting}
\label{sec: experimental}
The specific details of the experimental setup are as follows: (1) The training regimen involves training each model for up to 100 epochs, which refers to the maximum number of iterations over the entire training dataset. However, to prevent overfitting and unnecessary computation, we use an early stopping strategy. Early stopping monitors the model's performance on a validation set during training and halts the training process if there is no improvement in the validation performance over a specified number of consecutive epochs. This is done to ensure the model does not continue training unnecessarily once it has converged, saving computational resources. (2) The patience parameter is set to 10, meaning that if the validation performance does not improve for 10 consecutive epochs, training is stopped early. This parameter helps balance between giving the model enough time to learn and preventing overtraining when the model has already reached an optimal performance plateau.

 \section{Detailed experimental results}
In order to verify the superiority of TGFormer, we performed further analyses in Table~\ref{tab: tap} As shown in Table~\ref{tab: p}, each dataset is displayed in two rows: the first reports the 95\% CI and the second provides the p-value for the comparison of the TGFormer model with the baseline method. For TGFormer itself, we denote the p-value with a ``-'' because it is the reference value. The results show that TGFormer significantly outperforms the baseline method (p $<$ 0.05), confirming the statistical reliability of our findings.

\begin{table*}[!ht]
    \caption{95\% CI and p-values for transductive temporal link prediction with random, historical, and inductive negative sampling strategies. NSS is the abbreviation of Negative Sampling Strategies}
    \centering
      \resizebox{\linewidth}{!}
      {%
      \begin{tabular}{c|cccccccccccc}
    \midrule
        {NSS} & {Datasets} & JODIE        & DyRep        & TGAT         & TGN                   & CAWN   & EdgeBank   & TCL          & GraphMixer   & DyGFormer    &FreeDyG      & TGFormer \\
        \midrule
        \multirow{12}*{\rotatebox{90}{Random}}
        & Wikipedia  & \makecell[c]{[96.32, 96.68],\\ $<$0.001} & \makecell[c]{[94.78, 94.94],\\ $<$0.001} & \makecell[c]{[96.86, 97.02],\\ $<$0.001} & \makecell[c]{[98.37, 98.53],\\ $<$0.001} & \makecell[c]{[98.72, 98.80],\\ $<$0.001} & \makecell[c]{[90.37, 90.37],\\ $<$0.001} & \makecell[c]{[96.27, 96.67],\\ $<$0.001} & \makecell[c]{[97.21, 97.29],\\ $<$0.001} & \makecell[c]{[99.00, 99.06],\\ $<$0.001} & \makecell[c]{[99.25, 99.27],\\ $<$0.001} & \makecell[c]{[99.67, 99.91],\\ -}  \\
        & Reddit     & \makecell[c]{[98.13, 98.49],\\ $<$0.001} & \makecell[c]{[98.17, 98.27],\\ $<$0.001} & \makecell[c]{[98.49, 98.55],\\ $<$0.001} & \makecell[c]{[98.55, 98.71],\\ $<$0.001} & \makecell[c]{[99.10, 99.12],\\ $<$0.001} & \makecell[c]{[94.86, 94.86],\\ $<$0.001} & \makecell[c]{[97.50, 97.56],\\ $<$0.001} & \makecell[c]{[97.30, 97.32],\\ $<$0.001} & \makecell[c]{[99.21, 99.23],\\ $<$0.001} & \makecell[c]{[99.47, 99.49],\\ $<$0.001} & \makecell[c]{[99.84, 99.88],\\ -}  \\
        & LastFM     & \makecell[c]{[68.20, 73.50],\\ $<$0.001} & \makecell[c]{[69.18, 74.66],\\ $<$0.001} & \makecell[c]{[73.16, 73.68],\\ $<$0.001} & \makecell[c]{[71.56, 82.58],\\ $<$0.001} & \makecell[c]{[86.92, 87.06],\\ $<$0.001} & \makecell[c]{[79.29, 79.29],\\ $<$0.001} & \makecell[c]{[64.58, 69.96],\\ $<$0.001} & \makecell[c]{[75.28, 75.94],\\ $<$0.001} & \makecell[c]{[92.83, 93.17],\\ $<$0.001} & \makecell[c]{[91.93, 92.37],\\ $<$0.001} & \makecell[c]{[96.18, 97.56],\\ -}  \\
        & Enron      & \makecell[c]{[84.35, 85.19],\\ $<$0.001} & \makecell[c]{[77.71, 87.05],\\ $<$0.001} & \makecell[c]{[69.82, 72.42],\\ $<$0.001} & \makecell[c]{[84.99, 88.07],\\ $<$0.001} & \makecell[c]{[89.44, 89.68],\\ $<$0.001} & \makecell[c]{[83.53, 83.53],\\ $<$0.001} & \makecell[c]{[78.71, 80.69],\\ $<$0.001} & \makecell[c]{[82.03, 82.47],\\ $<$0.001} & \makecell[c]{[92.30, 92.64],\\ $<$0.001} & \makecell[c]{[92.44, 92.58],\\ $<$0.001} & \makecell[c]{[96.45, 97.83],\\ -}  \\
        & UCI        & \makecell[c]{[87.97, 90.89],\\ $<$0.001} & \makecell[c]{[61.93, 68.35],\\ $<$0.001} & \makecell[c]{[78.66, 80.60],\\ $<$0.001} & \makecell[c]{[90.89, 93.79],\\ $<$0.001} & \makecell[c]{[95.10, 95.26],\\ $<$0.001} & \makecell[c]{[76.20, 76.20],\\ $<$0.001} & \makecell[c]{[87.31, 91.83],\\ $<$0.001} & \makecell[c]{[92.46, 94.04],\\ $<$0.001} & \makecell[c]{[95.55, 96.03],\\ $<$0.001} & \makecell[c]{[96.13, 96.43],\\ $<$0.001} & \makecell[c]{[98.81, 99.37],\\ -}  \\
        & CollegeMsg & \makecell[c]{[72.87, 77.95],\\ $<$0.001} & \makecell[c]{[49.43, 64.41],\\ $<$0.001} & \makecell[c]{[79.87, 80.67],\\ $<$0.001} & \makecell[c]{[91.24, 94.00],\\ $<$0.001} & \makecell[c]{[95.78, 95.94],\\ $<$0.001} & \makecell[c]{[76.42, 76.42],\\ $<$0.001} & \makecell[c]{[83.50, 83.78],\\ $<$0.001} & \makecell[c]{[92.64, 93.44],\\ $<$0.001} & \makecell[c]{[95.76, 95.82],\\ $<$0.001} & \makecell[c]{[95.51, 96.11],\\ $<$0.001} & \makecell[c]{[98.89, 99.45],\\ -}  \\ \midrule
        \multirow{12}*{\rotatebox{90}{Historical}}
        & Wikipedia  & \makecell[c]{[82.10, 83.92],\\ $<$0.001} & \makecell[c]{[79.15, 80.71],\\ $<$0.001} & \makecell[c]{[87.07, 87.69],\\ $<$0.001} & \makecell[c]{[86.40, 87.32],\\ $<$0.001} & \makecell[c]{[68.88, 73.54],\\ $<$0.001} & \makecell[c]{[73.35, 73.35],\\ $<$0.001} & \makecell[c]{[88.51, 89.59],\\ $<$0.001} & \makecell[c]{[90.76, 91.04],\\ $<$0.001} & \makecell[c]{[78.69, 85.77],\\ $<$0.001} & \makecell[c]{[90.80, 92.38],\\ $<$0.001} & \makecell[c]{[92.23, 92.71],\\ -} \\
        & Reddit     & \makecell[c]{[79.53, 80.53],\\ $<$0.001} & \makecell[c]{[79.40, 80.26],\\ $<$0.001} & \makecell[c]{[79.27, 79.83],\\ $<$0.001} & \makecell[c]{[80.37, 82.07],\\ $<$0.001} & \makecell[c]{[80.19, 81.45],\\ $<$0.001} & \makecell[c]{[73.59, 73.59],\\ $<$0.001} & \makecell[c]{[76.92, 77.36],\\ $<$0.001} & \makecell[c]{[78.19, 78.69],\\ $<$0.001} & \makecell[c]{[80.64, 82.50],\\ $<$0.001} & \makecell[c]{[84.26, 87.08],\\ $<$0.001} & \makecell[c]{[82.58, 86.30],\\ -} \\
        & LastFM     & \makecell[c]{[69.05, 79.65],\\ 0.002          } & \makecell[c]{[71.50, 78.34],\\ 0.004           } & \makecell[c]{[71.26, 71.92],\\ $<$0.001} & \makecell[c]{[70.41, 83.33],\\ 0.013           } & \makecell[c]{[69.26, 70.46],\\ $<$0.001} & \makecell[c]{[73.03, 73.03],\\ $<$0.001} & \makecell[c]{[56.09, 62.51],\\ $<$0.001} & \makecell[c]{[71.79, 73.15],\\ $<$0.001} & \makecell[c]{[80.90, 82.24],\\ $<$0.001} & \makecell[c]{[79.00, 80.42],\\ $<$0.001} & \makecell[c]{[83.69, 85.35],\\ -} \\
        & Enron      & \makecell[c]{[66.09, 73.61],\\ $<$0.001} & \makecell[c]{[67.33, 75.05],\\ $<$0.001} & \makecell[c]{[62.61, 65.53],\\ $<$0.001} & \makecell[c]{[71.46, 76.36],\\ $<$0.001} & \makecell[c]{[64.23, 65.23],\\ $<$0.001} & \makecell[c]{[76.53, 76.53],\\ $<$0.001} & \makecell[c]{[70.12, 71.20],\\ $<$0.001} & \makecell[c]{[76.70, 79.26],\\ $<$0.001} & \makecell[c]{[74.62, 76.64],\\ $<$0.001} & \makecell[c]{[77.73, 80.01],\\ $<$0.001} & \makecell[c]{[80.92, 84.36],\\ -} \\
        & UCI        & \makecell[c]{[67.17, 83.31],\\ 0.018           } & \makecell[c]{[50.73, 59.47],\\ $<$0.001} & \makecell[c]{[66.36, 70.18],\\ $<$0.001} & \makecell[c]{[77.48, 83.38],\\ 0.002           } & \makecell[c]{[64.69, 65.91],\\ $<$0.001} & \makecell[c]{[65.50, 65.50],\\ $<$0.001} & \makecell[c]{[76.44, 84.06],\\ 0.003           } & \makecell[c]{[82.21, 86.01],\\ 0.045           } & \makecell[c]{[81.03, 83.31],\\ $<$0.001} & \makecell[c]{[84.44, 87.767],\\ 0.04           } & \makecell[c]{[84.26, 87.98],\\ -} \\
        & CollegeMsg & \makecell[c]{[55.34, 73.48],\\ 0.013           } & \makecell[c]{[46.37, 49.09],\\ $<$0.001} & \makecell[c]{[66.74, 69.62],\\ $<$0.001} & \makecell[c]{[79.08, 82.22],\\ $<$0.001} & \makecell[c]{[84.39, 84.69],\\ $<$0.001} & \makecell[c]{[44.16, 44.16],\\ $<$0.001} & \makecell[c]{[68.42, 68.64],\\ $<$0.001} & \makecell[c]{[83.81, 84.05],\\ $<$0.001} & \makecell[c]{[80.42, 81.44],\\ $<$0.001} & \makecell[c]{[88.64, 89.58],\\ $<$0.001} & \makecell[c]{[88.58, 91.46],\\ -} \\ \midrule
        \multirow{12}*{\rotatebox{90}{Inductive}}
        & Wikipedia  & \makecell[c]{[74.61, 76.69],\\ $<$0.001} & \makecell[c]{[68.01, 72.41],\\ $<$0.001} & \makecell[c]{[86.78, 87.22],\\ $<$0.001} & \makecell[c]{[85.01, 86.23],\\ $<$0.001} & \makecell[c]{[70.41, 77.71],\\ $<$0.001} & \makecell[c]{[80.63, 80.63],\\ $<$0.001} & \makecell[c]{[85.76, 87.76],\\ $<$0.001} & \makecell[c]{[88.35, 88.83],\\ $<$0.001} & \makecell[c]{[70.79, 85.79],\\ $<$0.001} & \makecell[c]{[89.06, 91.04],\\ $<$0.001} & \makecell[c]{[92.22, 93.98],\\ -} \\
        & Reddit     & \makecell[c]{[86.76, 87.20],\\ $<$0.001} & \makecell[c]{[85.94, 86.66],\\ $<$0.001} & \makecell[c]{[89.26, 89.92],\\ $<$0.001} & \makecell[c]{[87.77, 88.43],\\ $<$0.001} & \makecell[c]{[91.34, 92.00],\\ $<$0.001} & \makecell[c]{[85.48, 85.48],\\ $<$0.001} & \makecell[c]{[87.05, 87.85],\\ $<$0.001} & \makecell[c]{[85.11, 85.41],\\ $<$0.001} & \makecell[c]{[90.55, 91.67],\\ $<$0.001} & \makecell[c]{[90.50, 90.98],\\ $<$0.001} & \makecell[c]{[90.32, 94.42],\\ -} \\
        & LastFM     & \makecell[c]{[56.45, 68.89],\\ 0.046           } & \makecell[c]{[60.65, 68.17],\\ 0.031           } & \makecell[c]{[70.93, 71.33],\\ 0.003           } & \makecell[c]{[54.64, 77.26],\\ 0.323           } & \makecell[c]{[66.41, 68.55],\\ 0.011           } & \makecell[c]{[75.49, 75.49],\\ 0.999           } & \makecell[c]{[57.07, 59.35],\\ $<$0.001} & \makecell[c]{[67.66, 68.58],\\ 0.002           } & \makecell[c]{[73.30, 74.64],\\ 0.001           } & \makecell[c]{[71.89, 72.49],\\ 0.001           } & \makecell[c]{[72.34, 75.68],\\ -} \\
        & Enron      & \makecell[c]{[67.60, 70.32],\\ $<$0.001} & \makecell[c]{[65.67, 69.91],\\ $<$0.001} & \makecell[c]{[62.03, 65.85],\\ $<$0.001} & \makecell[c]{[67.09, 74.69],\\ $<$0.001} & \makecell[c]{[74.39, 75.91],\\ $<$0.001} & \makecell[c]{[73.89, 73.89],\\ $<$0.001} & \makecell[c]{[70.88, 71.70],\\ $<$0.001} & \makecell[c]{[73.90, 76.12],\\ $<$0.001} & \makecell[c]{[76.17, 78.65],\\ $<$0.001} & \makecell[c]{[76.92, 78.70],\\ $<$0.001} & \makecell[c]{[82.25, 86.43],\\ -} \\
        & UCI        & \makecell[c]{[64.14, 67.84],\\ $<$0.001} & \makecell[c]{[52.32, 57.26],\\ $<$0.001} & \makecell[c]{[67.50, 69.84],\\ $<$0.001} & \makecell[c]{[69.95, 71.93],\\ $<$0.001} & \makecell[c]{[63.94, 65.28],\\ $<$0.001} & \makecell[c]{[57.43, 57.43],\\ $<$0.001} & \makecell[c]{[74.47, 77.55],\\ $<$0.001} & \makecell[c]{[79.40, 80.80],\\ $<$0.001} & \makecell[c]{[69.87, 74.63],\\ $<$0.001} & \makecell[c]{[81.34, 83.36],\\ $<$0.001} & \makecell[c]{[79.26, 82.14],\\ -} \\
        & CollegeMsg & \makecell[c]{[52.75, 54.23],\\ $<$0.001} & \makecell[c]{[51.86, 56.40],\\ $<$0.001} & \makecell[c]{[67.36, 69.72],\\ $<$0.001} & \makecell[c]{[71.82, 71.98],\\ $<$0.001} & \makecell[c]{[75.00, 75.24],\\ $<$0.001} & \makecell[c]{[43.49, 43.49],\\ $<$0.001} & \makecell[c]{[68.67, 68.73],\\ $<$0.001} & \makecell[c]{[79.05, 79.29],\\ $<$0.001} & \makecell[c]{[69.28, 71.60],\\ $<$0.001} & \makecell[c]{[79.82, 83.02],\\ $<$0.001} & \makecell[c]{[77.98, 82.32],\\ -} \\ \midrule
    \end{tabular}
    }
\label{tab: p}
\end{table*}










\end{document}